\documentclass{article}

\usepackage{graphicx} 
\usepackage{subcaption}
\usepackage{amsmath}
    \usepackage[preprint]{neurips_2025}

\usepackage[utf8]{inputenc} % allow utf-8 input
\usepackage[T1]{fontenc}    % use 8-bit T1 fonts
\usepackage{hyperref}       % hyperlinks
\usepackage{url}            % simple URL typesetting
\usepackage{booktabs}       % professional-quality tables
\usepackage{amsfonts}       % blackboard math symbols
\usepackage{nicefrac}       % compact symbols for 1/2, etc.
\usepackage{microtype}      % microtypography
\usepackage{xcolor}         % colors
\usepackage{graphicx}
\usepackage{amsmath}
\usepackage{caption}
\usepackage{booktabs}
\usepackage{hyperref}
\usepackage{enumitem}

\title{From Local Cues to Global Percepts: \\Emergent Gestalt Organization in \\Self-Supervised Vision Models}

\author{%
  Tianqin Li, ~ Ziqi Wen, ~ Leiran Song, ~ Jun Liu, ~ Zhi Jing, ~ Tai Sing Lee \\\\
  Carnegie Mellon University \\\\
  \texttt{\{tianqinl, ziqiw, leirans, junl, zhij, taislee\}@andrew.cmu.edu}
}

\begin{document}

\maketitle

\begin{abstract}
Human visual perception organizes local cues into coherent global forms using Gestalt principles such as closure, proximity, and figure–ground assignment—functions that rely on sensitivity to global spatial structure. In this work, we investigate whether modern vision models exhibit similar perceptual behaviors, and under what training conditions such behaviors emerge. We first show that Vision Transformers (ViTs) trained with Masked Autoencoding (MAE) display internal activation patterns consistent with Gestalt laws, including illusory contour completion, convexity preference, and dynamic figure–ground segregation. To better understand the computational basis of this behavior, we hypothesize that modeling global visual dependencies is a necessary condition for Gestalt-like organization to arise. To evaluate this capability directly, we introduce the Distorted Spatial Relationship Testbench (DiSRT)—a benchmark that tests a model’s sensitivity to global spatial relationship perturbations while preserving local textures. Using DiSRT, we find that self-supervised models (e.g., MAE, CLIP) outperform supervised baselines and, in some cases, even exceed human performance. 
Interestingly, ConvNeXt models trained with MAE also exhibit Gestalt-compatible representations, suggesting that global structure sensitivity can emerge independently of architectural inductive biases. Yet this capability proves fragile: standard classification finetuning substantially degrades a model's DiSRT performance, indicating that supervised objectives may suppress global perceptual organization. Inspired by biological vision, we show that a simple Top-K activation sparsity mechanism can effectively restore global sensitivity in these models.
Altogether, our results identify the training conditions that promote or suppress Gestalt-compatible perception in deep models, and establish DiSRT as a principled diagnostic for global perceptual sensitivity acoss architectures and training objectives. 

\end{abstract}

\section{Introduction}

% In this paper, we introduce sparse MAE for 

% From david marr vision book: Figure 2-6 (section 2.1 physical background of early vision). "Subjective countours. The visual system apparently regards changes in depth as so important that they must be made explicit everywhere, including places where there is no direct visual evidence for them. "

% This is a discription of Kanisza Triangle where the depth is made explicit by our visual system. 

% In this paper, we want to connect message with depth information. 

Human visual perception is holistic~\cite{wagemans2012centuryI, wagemans2012centuryII, wertheimer1923}: rather than interpreting scenes as collections of isolated elements, the brain spontaneously organizes them into coherent structures. This process is governed by a set of well-established heuristics known as the Gestalt principles—such as closure, proximity, continuity, and figure-ground organization—which describe how global perceptual structure arises from the spatial configuration of local cues.

Importantly, these principles depend on the global relationships among visual elements. For example, perceiving a Kanizsa triangle requires integrating disconnected edge segments into an illusory shape~\citep{kanizsa1976subjective}; identifying a figure in front of a background relies on the comparison of convexity and border ownership across an extended spatial field~\citep{wagemans2012centuryI}. In this sense, Gestalt perception is not merely a psychological curiosity—it is a behavioral indicator of how the visual system processes long-range dependencies across space. If a system exhibits Gestalt-like behaviors, it must encode information that spans beyond local receptive fields, capturing the joint configuration of distributed parts of the image.

This insight raises a key question in modern computer vision: Do neural networks trained on natural images also develop Gestalt-like global sensitivities—and if so, under what conditions? While it has been suggested that Vision Transformers (ViTs), by virtue of self-attention, are well-suited to modeling long-range dependencies, the empirical evidence for their alignment with Gestalt perception remains limited. Prior work on “shape bias” and robustness has typically used style transfer or occlusion-based perturbations, which conflate texture insensitivity with structural understanding.

In this paper, we provide a direct investigation of Gestalt perception in neural networks, using both classic visual illusions and quantitative benchmarks. We show that ViTs trained with Masked Autoencoding (MAE) exhibit internal representations aligned with Gestalt laws—e.g., forming coherent figure-ground separation, grouping elements by continuity and proximity, and preferring convex figures. We interpret this behavior as evidence that MAE encourages models to develop global perceptual integration capabilities, not merely through architectural inductive bias, but through the training objective itself.

To evaluate this hypothesis more systematically, we introduce the Distorted Spatial Relationship Testbench (DiSRT), a new benchmark that assesses a model’s sensitivity to global structural perturbations while preserving local texture statistics. DiSRT isolates long-range visual dependencies by requiring models (or humans) to detect spatial relationship perturbation generated via texture synthesis. Through this benchmark, we demonstrate that:

\begin{itemize}
    \item Self-supervised models like MAE and CLIP dramatically outperform supervised baselines on DiSRT.
    \item ConvNets, when trained with MAE, also develop Gestalt-like representations.
    \item Standard classification finetuning erodes these global sensitivities, but can be partially recovered through a biologically inspired Top-K activation sparsity mechanism.
\end{itemize}

Our results suggest that Gestalt perception can emerge naturally in modern vision systems, but is critically dependent on training objectives that promote global context modeling. These findings offer a bridge between classic theories of perception and current advances in self-supervised learning, providing new insights into the development of structured representations in neural networks.

\section{Gestalt Perception in Mask Autoencoders}

\begin{figure}[htbp]
    \centering
    \includegraphics[width=0.8\linewidth]{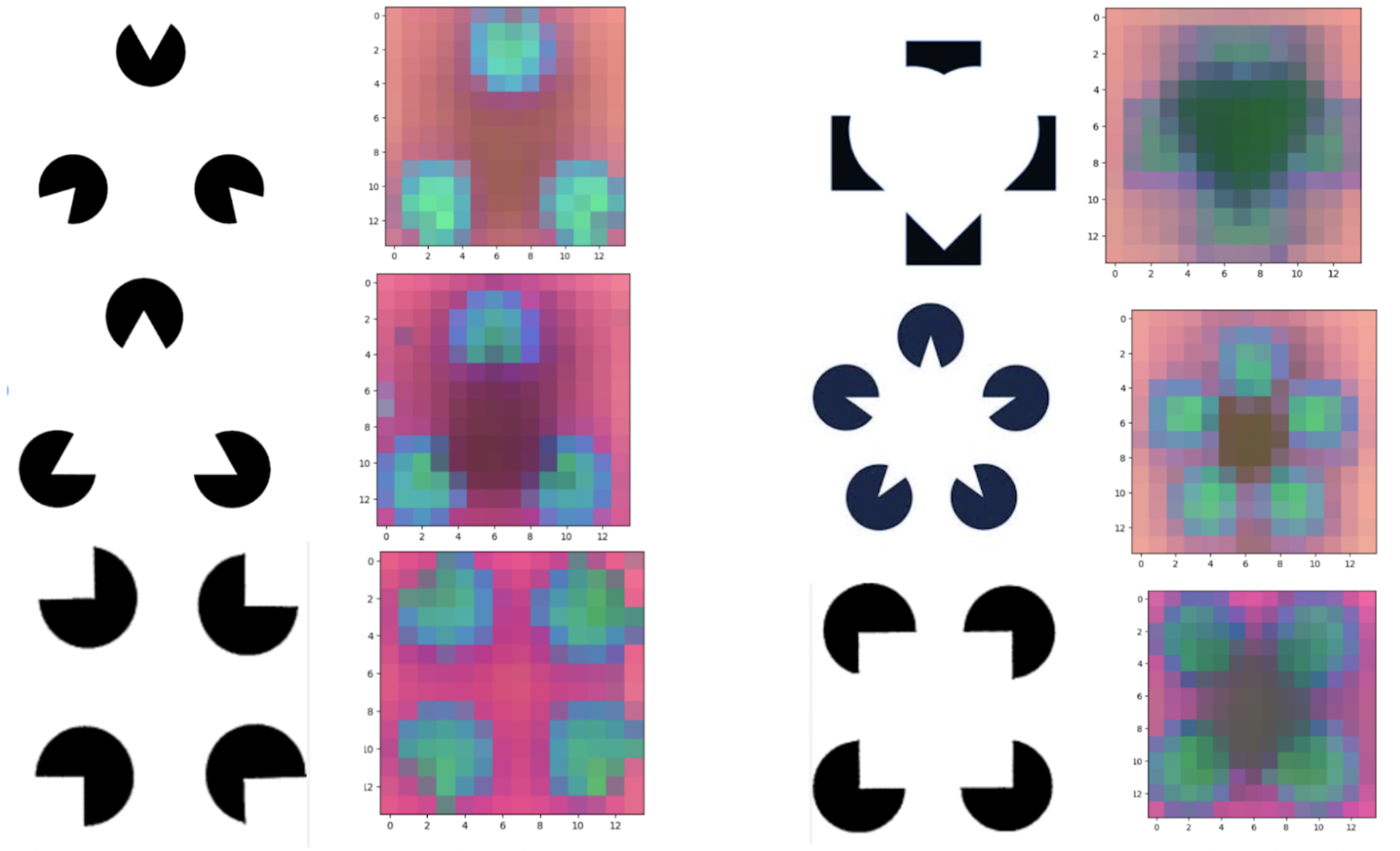}
    \caption{Internal activation of ViT Mask Autoencoder on several illustory countour images. We project the internal activation of certain attention head to 3 dimension space using PCA and color the projection with RGB colors. The PCA basis is shared across different images. We can see that (1) the PCA clearly captures the direction that seperates the figure and the ground. (2) The Kanisza Trinagle is perceived when the illustory countour is aligned and not seen when the pac-man are not very well coordinated, demonstrating the ability of MAE trained ViT in Gestalt Perception. Please refer to the supplementary materials for method used here.}
    \label{fig:ktriangle}
\end{figure}

\paragraph{Figure-Ground: The Kanisza Triangles Illusion.}
The Kanizsa Triangle illusion is a classic example of how our visual system perceives global structures through the Gestalt principle of closure. In this illusion, three 'pac-man' shapes are arranged in such a way that the mind perceives an equilateral triangle, even though no triangle is physically drawn. This illusion highlights the brain's ability to integrate local elements into a coherent whole, demonstrating the importance of global structure in visual perception. However, it's not been clearly shown before that neural networks learned with natural image training also can see the Kanisza  Illusion. Utilizing ViT models trained with Mask Autoencoder Objective~\cite{he2022masked}, we demonstrate that certain attention head in the ViT layers exhibits strong figure ground seperation capability -- an ability that requires integrating the local visual cue into global perception as shown in Figure~\ref{fig:ktriangle} (See caption for details).

\begin{figure}[htbp]
    \centering
    \includegraphics[width=0.9\linewidth]{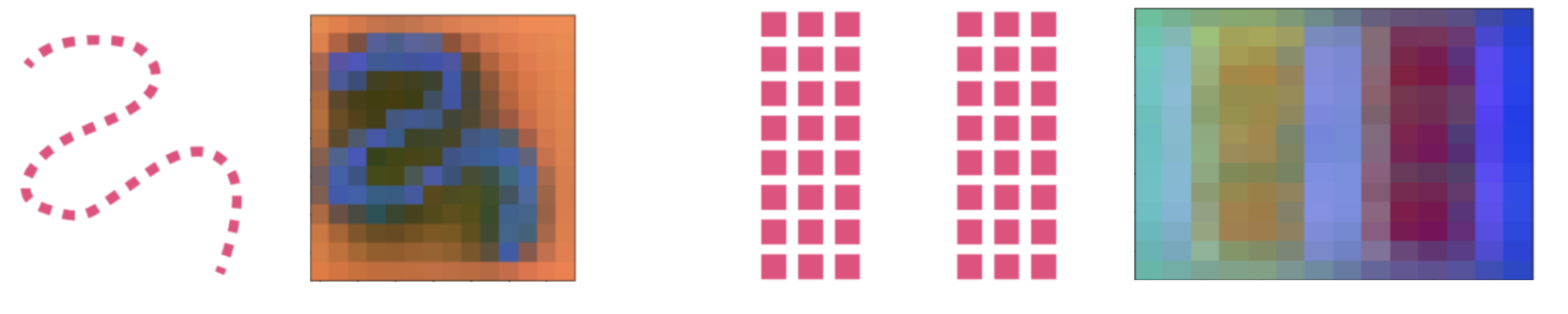}
    \caption{Other Gestalt Perception Rules learned in ViT-MAE: Law of Continuity and Proximity}
    \label{fig:other-gestalt}
\end{figure}

\paragraph{Law of Continuity and Proximity.}In addition to demonstrating figure-ground separation, our analysis also reveals that ViT-MAE exhibit perceptual behaviors aligned with other Gestalt principles, notably  \textit{Law of Continuity} and \textit{Law of Proximity}. The law of continuity posits that discrete elements aligned along a continuous line or curve are perceived as related. As shown in the Figure~\ref{fig:other-gestalt} Left, activations along the contour are much similar in activation regardless whether the pixel contains a dot or blank background. The entire curve is perceived as a whole rather than discretized composing dots. The model can also see law of proximity. On the right of Figure~\ref{fig:other-gestalt}, when groups of squares are spatially proximate, the ViT-MAE assigns similar activation values to these proximate objects (similar PCA projection color). We can see a clear grouping effect where two proximate groups have different activation values. These two additional laws further hint that ViT trained by MAE strategy could perceive global structure of the image exceptionally well, aligning with gestalt theory of human perception. 

\begin{figure}
    \centering
    \includegraphics[width=0.9\linewidth]{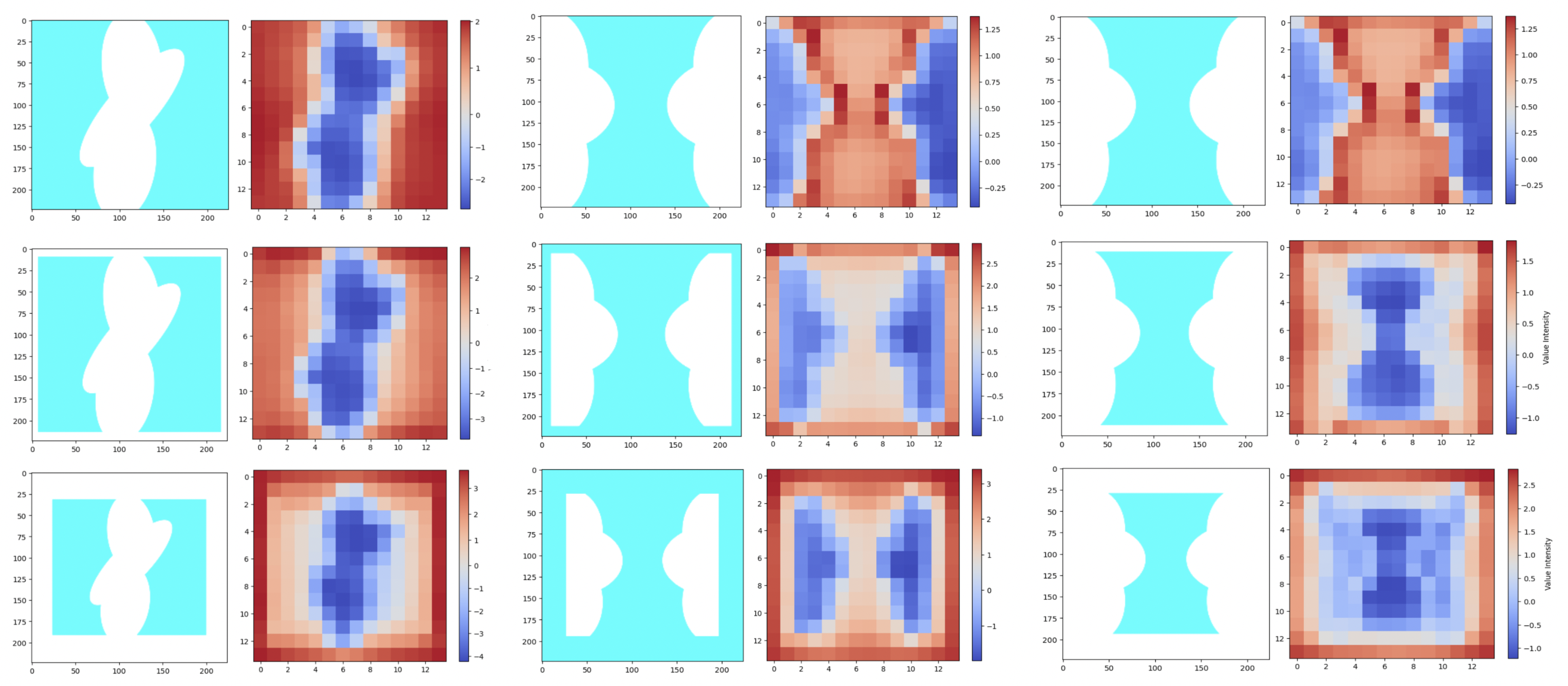}
    \caption{Law of Convexity. Convex and Concave preference for figure ground identification. \textbf{First Row}: Law of Convexity. The Model prefer the convex shape as the figure in the figure ground relationship. As shown in the supplementary, blue / negative PCA projection value is associated with the figure and the the orange / postive value is associated with the ground. The results suggest that model prefers the convex object as the figure and the concave shape as the background. \textbf{Second and Third Row}: Adding margins to dynamically manipulate the figure ground relationship. In the first two columns, adding border won't change the figure ground relationshp, the convex object still wins. However, in the third column, adding border would flip the figure ground relationship, indicating that the model is capable of performing dynamic spatial reasoning integrating global perceptual cues.}
    \vspace{-2mm}
    \label{fig:convexity}
\end{figure}

\paragraph{Law of Convexity.} Human psychology experiments suggest that humans prefer convex object as figure and concave visual stimulus as background~\cite{arnheim1972art, kanizsa1976convexity, peterson2008inhibitory, bertamini2006owns}. However, whether this is a universal principle for all vision system or specifically tune to human is not clear. Here, we find that ViT trained by MAE can also exhibit such convexity preference for figure. We demonstrate this by first constructing cue-conflict images following similar idea in human psychology study~\cite{kanizsa1976convexity, peterson2008inhibitory} and analyze the internal activation patterns of ViT-MAE. In Figure~\ref{fig:convexity}, a convex and a concave pattern are presented to the network. The model's internal representation, summarized by the first principle component projection the internal activation, is shown on the right of each image. The PCA basis is the share across all images. The value of the PCA projection (positive and negative) is associated with the figure ground assignment -- as we show in the supplementary that in natural images, figure always results in negative value of such PCA projection and ground results in positive value. Hence, by looking at the value of the PCA projection, we can sense how the model perceives the figure-ground organization visually. 
Utilizing the PCA values, we see the law of convexity from Figure~\ref{fig:convexity}: If we look at the \textbf{first row} of Figure~\ref{fig:convexity}, when a convex and concave cue both present in the image, the model considers the convex cue (the half face and the man in the center) as figure object (blue color / low value of PCA projection) and the concave cue (e.g. the vass) as background (orange color / high value in PCA projection). This indicates that the ViT-MAE model learns the law of convexity, aligning with human perception.

\paragraph{Dynamic Spatial Reasoning of Gestalt Principle.} Interestingly, we further test the models' capability to perform spatial reasoning by adding different margins ( \textbf{second} and \textbf{third} row). In the first and second column, adding margins did not change the figure ground organization, however in the third column, the figure ground assignmet flips. These interesting results suggest that the model can reason the figure ground dynamically based on the other parts of the image, indicating an advanced spatial reasoning capability is learned. Notably, such behavior is consistent with human perceptual tendencies.

% Speficially, if we first look at the \textbf{first row} of Figure~\ref{fig:convexity}, when a convex and concave cue both present in the image, the model considers the convex cue as figure object (blue color / low value of PCA projection) and the concave cue (e.g. the vass) as background (orange color / high value in PCA projection). When adding the color of center object to the margin (second row and third row), the convex object remains the figure whereas the concave shape remains the background, suggesting a strong preference for convexity as figure. 

% Interestingly, if one add more and more white margin to the images (second row and third row of Figure~\ref{fig:convexity}), the model would start to see the vass, eventually consider the vass as an figure object. This further indicates the complex capability of the ViT-MAE to integrate long range global image cue together to produce human-like visual perception.

\paragraph{Is Gestalt Perception Exclusive to ViTs?}

With the popularity of the Vision Transformers in 2020s, it is natually to think the amazing capability of Gestalt perception of ViT-MAE largely come from the transformer architecture where the global relationship is modeled via the computational intensive but powerful self-attention mechanism. However, in this section, we demonstrate that the Gestalt Rule observed in previous section in ViT-MAE can also be found in convoluitonal neural networks. This further ponders the question of what contributes to the success of long range visual dependency modeling.

\begin{figure}[htbp]
  \centering
  \begin{subfigure}[b]{0.3\textwidth}
    \centering
    \includegraphics[width=\textwidth]{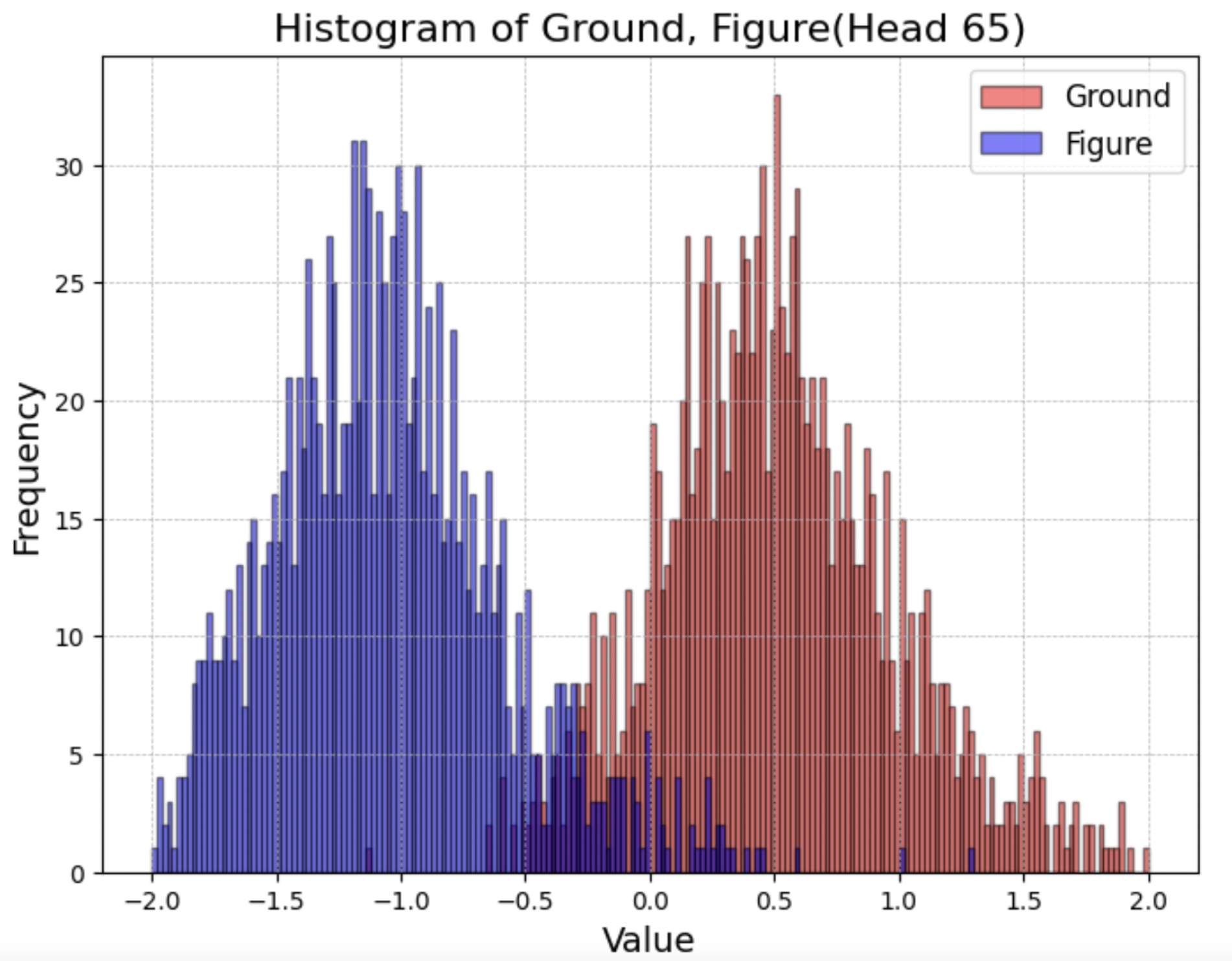}
    \caption{ViT-MAE's activation}
    \label{fig:subfig1}
  \end{subfigure}
  % \hfill
  \begin{subfigure}[b]{0.31\textwidth}
    \centering
    \includegraphics[width=\textwidth]{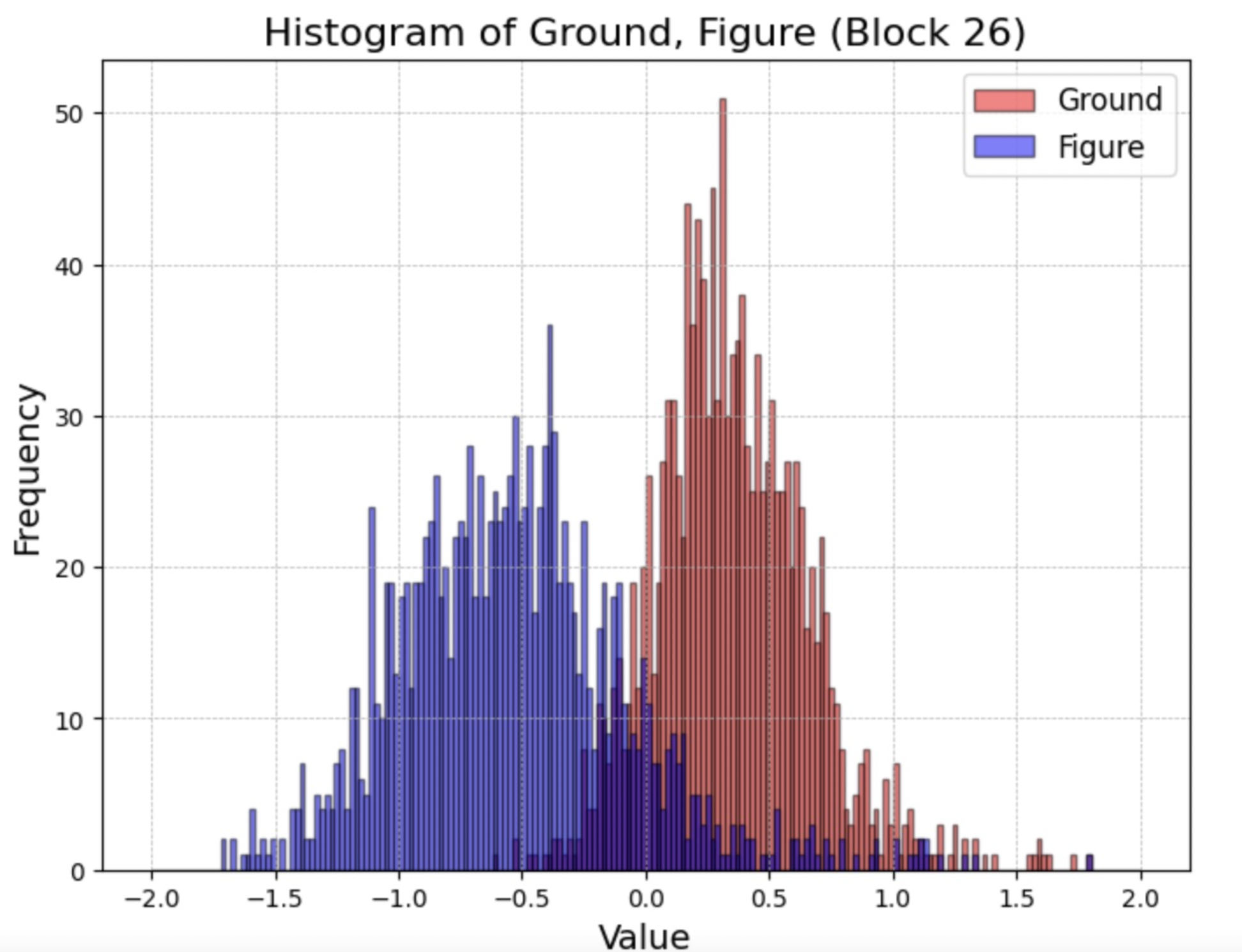}
    \caption{ConvNeXt-MAE's activation}
    \label{fig:subfig2}
  \end{subfigure}
  % \begin{subfigure}[b]{0.32\textwidth}
  %   \centering
  %   \includegraphics[width=\textwidth]{images/fg_seperation_convnext_v1.png}
  %   \caption{ConvNeXt-v1's activation}
  %   \label{fig:subfig2}
  % \end{subfigure}
  \caption{We compute the first principal component (PC1) over all activation vectors collected from a fixed layer across the entire set of 1,495 PASCAL VOC images. This shared PCA basis is then used to project each pixel’s activation onto PC1, yielding a scalar map per image. We label each pixel as figure or ground using the dataset's ground-truth segmentation masks and plot the resulting distribution of PC1 projection values. Both ViT-MAE (left) and ConvNeXt-MAE (right) show clear separability between figure (blue) and ground (orange) pixels in this shared feature space. This suggests that these self-supervised models encode figure-ground organization at the representation level—consistent with the Gestalt principle of figure-ground segregation—even without explicit supervision or segmentation objectives.
}
  \label{fig:fg_activations}
\end{figure}

Due to the limited page size, we briefly show that a convolutional neural network can also exhibit the Gestalt's figure-ground organization rule and leave the discussion of the rest of its gestalt perception behavior in the supplementary. Specifically, we analysize a state-of-the-art convolutional neural network family -- the ConvNeXt-v1~\cite{liu2022convnet} and ConvNeXt-MAE~\cite{woo2023convnext} -- by their response to a subset of PASCAL Challenge Images (1495 samples, see details in supplementary). To clearly see the results, we gather the internal activation (e.g. an activation tensor of one layer of shape [C, H, W]) of these models and compute the PCA basis of the C dimensional space. We then take the first principle component and compute the activation projection to this basis, resulting in [1, H, W] tensor for projections. This is visualized in Figure~\ref{fig:convexity} as well as plotted as histogram in Figure~\ref{fig:fg_activations}. The only difference in Figure~\ref{fig:fg_activations} is that we aggregate across all subset of images of PASCAL images resulting in [N, 1, H, W] tensors, where N indicates the number images we used. We then flatten the resulting value to [N * H * W, ] and plot them as histgram in Figure~\ref{fig:fg_activations}. 

Since PASCAL images has segmentation labels serving as a ground truth for figure-ground segementation, we utilize this information and label the histogram with their corresponding ground truth labels. Specifically, the segmentation object is considered figure and the rest is considered as ground. The results in Figure~\ref{fig:fg_activations} suggests that (1) our previous finding about figure ground stands on the natural image setting, i.e. the model is capable of separating figure and ground in its internal activation; (2) The figure ground perception rule of Gestalt is not exclusive to the ViT architecture as ConNeXt-MAE also exhibit such figure-ground gestalt separability.

\section{From Perception to Computation: Linking Gestalt Organization with Global Dependency between Visual Elements}

The emergence of Gestalt-like behaviors in ViT-MAE models—such as figure-ground separation, closure, and convexity preference—raises a key question: What computational properties enable these perceptual capabilities to emerge?

Classic theories of vision, including Marr’s framework, suggest that perceptual organization depends on integrating local cues into coherent global structures. Gestalt principles like closure or proximity are not detectable through local filters alone—they require modeling long-range spatial dependencies. Thus, we hypothesize that Gestalt-like behavior in neural networks arises from a model’s ability to represent and utilize global structure in their computation.

To investigate this systematically, we introduce the Distorted Spatial Relationship Testbench (DiSRT). DiSRT evaluates whether a model can distinguish long range spatial relationship distortions while preserving local visual cues, thereby isolating the model’s sensitivity to spatial configuration. While DiSRT does not directly test Gestalt laws, it measures a necessary computational condition for them: the capacity to encode relationships across distant parts of an image.

In the next section, we discuss about the construction of DiSRT benchmark and then show the results of evaluating a range of models and training objectives on DiSRT. 

% This question is highly related to the first question as learning gestalt perception requires a model to be sensitive to the global structure of the image. For example, when adding the margin in Figure~\ref{fig:convexity}, the perception of the figure ground organization would change. If a model is not sensitive to the global image structure but rather perceive the image as a bag-of-words of local parts, the model would not sense such changes, i.e. the whole is greater than the sum of its part only when the local parts can influence and interact with each other. Hence answering the second question would contribute significantly to the broader puzzle of gestalt perception. In the remaining section of the paper, we systemaically study the later question by constructing a benchmark designed to test the model's capability of sensing long-range visual dependency. 

% \newpage
\section{DiSRT Benchmark Construction }
\begin{figure}[ht!]
     \centering
     % \hfill
     \begin{subfigure}[b]{0.45\textwidth}
         \centering
         \includegraphics[width=0.85\textwidth]{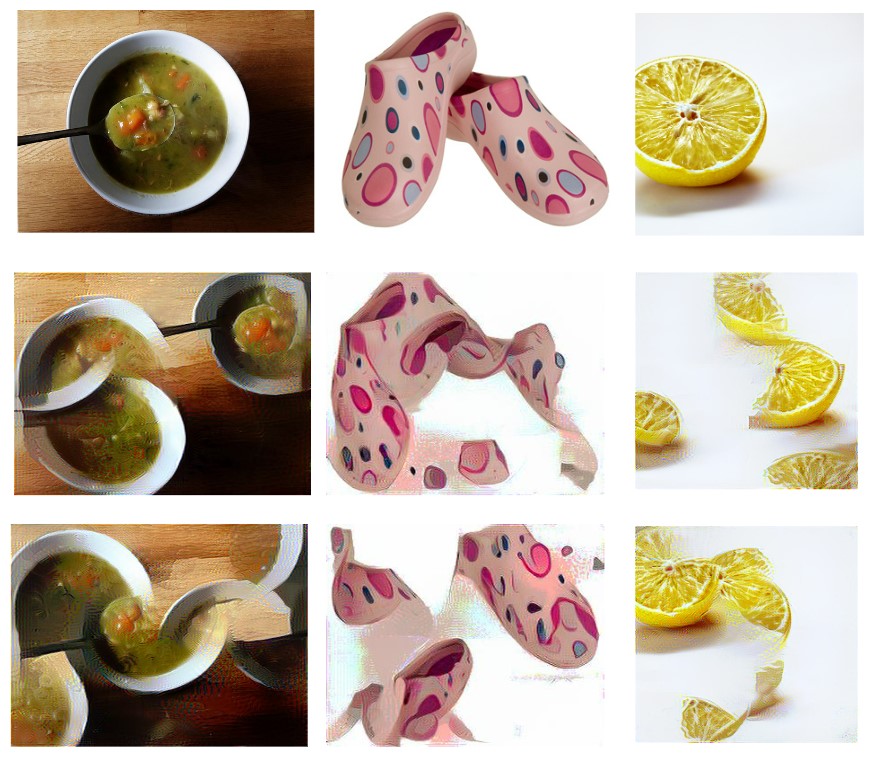} 
         \caption{Examples in DiSRT}
         \label{fig:DistExample}
     \end{subfigure}
     \begin{subfigure}[b]{0.45\textwidth}
         \centering
         \includegraphics[width=\textwidth]{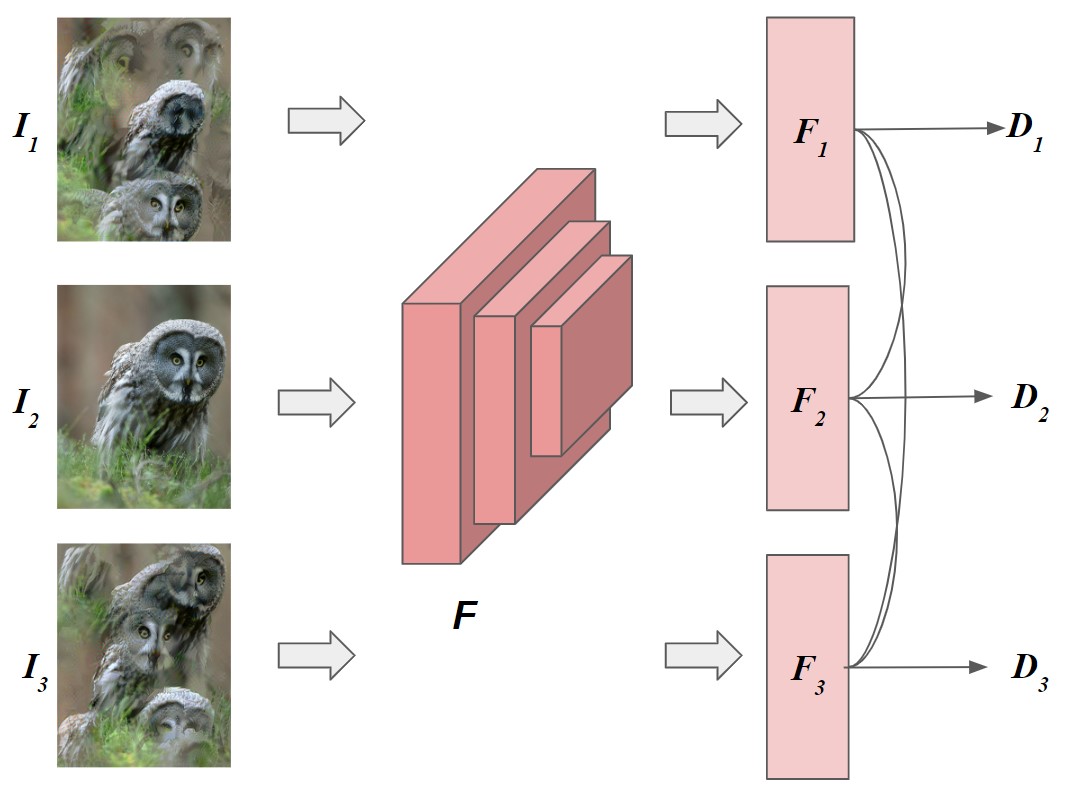}
         \caption{The process of DiSRT for models}
         \label{fig:DistProcess}
     \end{subfigure}
        \caption{Distorted Spatial Relationship Testbench (DiSRT)}
        \label{fig:Dist}
\end{figure}

Our Distorted Spatial Relationship Testbench (DiSRT) deliberately compares the representation before and after we apply the global spatial relationship distortion. One could imagine there could be many global distortion variants of an original image as the joint spatial configuration of local parts could be arbitrary. To obtain a quantitative measurement of the model's global spatial relationship sensitivity, DiSRT formulates the evaluation metrics as the accuracy of an oddity detection task. The subjects to DiSRT (machine learning models or human) are asked to select a distinct image from a pool of choices, which consists one original image where the global relationship is intact, and $N$ global spatial relationship distorted variants of the original image (each of which preserves the local patterns) . We pick $N=2$ for all the DiSRT test as we observe that increases $N$ will not increase the difficulty of the task.
\begin{figure}[h]
    \centering
    \includegraphics[width=0.75\linewidth]{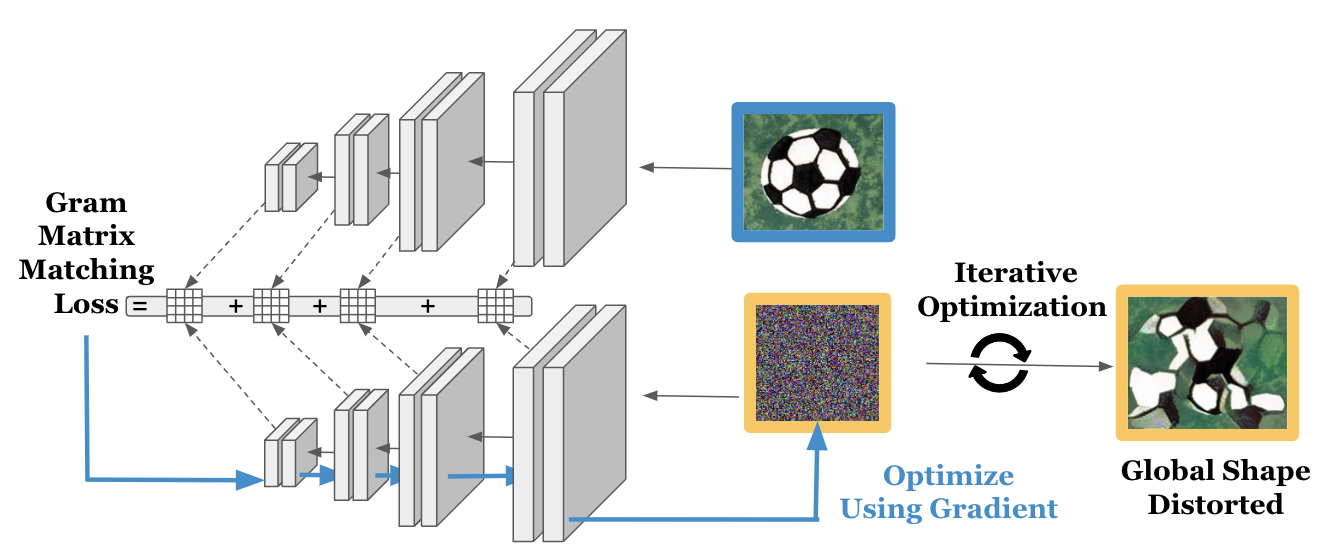}
    \caption{Mechanism of computing global relationship distorted images. We implement approach proposed in \cite{gatys2015texture}. Specifically, we optimize a randomly initialized image (yellow) so that when it passes through a pretrained VGG network, its intermediate layers' gram matrix match the targeted image (blue). This results in preserving the images' local features but randomizing the global structures.}
    \vspace{-2mm}
    \label{fig:texform}
\end{figure}

% \textcolor{red}{TODO: Explain the Texture Synthesis }
\vspace{-2mm}
\paragraph{Spatial Relationship Distortion via Texture Synthesis Program}

Texture Synthesis allows for the generation of images that retain the original texture details while randomizing the global structures. We utilize \cite{gatys2015texture} in particular to construct the spatial relationship distorted images for the DiSRT oddity detection task. We illustrate the process of this texture synthesis in Figure~\ref{fig:texform}. For any given target image, $I_t\in R^{(3, H_0, W_0)}$ (blue boundary image in Figure~\ref{fig:texform}), we want to get $I_o \in R^{(3, H_0, W_0)}$ that possess the same local features but distorted global spatial relationship (as shown in yellow boundary images on the right). To achieve this, we initialize tensor $I_1 \in R^{(3, H_0, W_0)}$ using value independently sampled from a isotropic Gaussian distribution and complete the process of $I_1 \rightarrow I_2 \rightarrow I_3 ... \rightarrow I_o$ through minimizing the \textit{L} as Gram Matrix Matching Loss. Specifically,
$$\frac{\partial Loss}{\partial I_i} = \frac{\partial }{\partial I_i}\sum_{l} || \text{Gram}(A_l(I_t)) - \text{Gram}(A_l(I_i))||^2$$
, where $A_l(I)\in R^{(C_l, H_l, W_l)}$ denotes the $l$-th layers activation tensor from which we destroy the global spatial information by computing the channel-wise dot products, i.e. $\text{Gram}(A_l(I_t))\in R^{(C_l, C_l)}$. 

\vspace{-3mm}
\paragraph{DiSRT Metric Formulation}For each trial in the Distorted Spatial Relationship Testbench (DiSRT), two spatial relationship distorted versions are generated using distinct random seeds. Each image, denoted as $I_i$, is then passed through the evaluation network $F$ to obtain a feature vector $F(I_i)$ from the final layer. The model identifies the image most dissimilar to the others by calculating the cosine distance between feature vectors. The procedure for this calculation is as follows:

\begin{equation}
D_i=\sum_{j\neq i}(1-\frac{F(I_i) \cdot F(I_j)}{\|F(I_i)\|_2\|F(I_j)\|_2}) / N
\end{equation}%\label{eq:}

$N$ represent the number of spatial organization distorted images in each traial, in DiSRT it would be equal to 2. The dissimilar of the image $I_i$ to other 2 images is calculated as the average of the pairwise cosine distance of each two image pairs. Cosine distance of vector $u$ and $v$ is calculated as $D_C(u,v)=1-S_C(u,v)$, where $S_C(u,v)$ is the cosine similarity. 

Model will select the images $a$ that is the most different from the other two images based on $D_i$:

\begin{equation}
a=\operatorname*{arg\,max}_i \frac{exp(D_i)}{\sum_jexp(D_j))}
\end{equation}%\label{eq:}

% \textcolor{red}{Part 1: Explain the how we create the distorted shape}
% \textcolor{red}{Part 2: Show the testing process}
\vspace{-2mm}
\subsection{Psychophysical Experiments}

Human vision is known to exhibit a strong bias towards perceiving the "gestalt" of the image, i.e. seeing the global structure by integrating the local cues. To quantify the gap between deep learning models and the human visual system, we conducted a psychophysical experiment with human subjects. To align this experiment closely with deep learning evaluations, participants were simply instructed to select the image they found to be "the most different," without receiving any additional hints or context. To mimic the feedforward processes in deep learning models, we displayed stimulus images for a limited time, thereby restricting additional reasoning. Furthermore, participants received no feedback on the correctness of their selections, eliminating the influence of supervised signals.

In each trial, participants were simultaneously presented with three stimulus images for a duration of 800 ms. They then had an additional 1,200 ms, making a total of 2,000 ms, to make their selections. Any response given after the 2,000-millisecond window was considered invalid. To mitigate the effects of fatigue, participants were allowed breaks after completing 100 trials, which consisted of 100 sets of images. We accumulated data from 16,800 trials and 32 human subjects to calculate the overall performance on DiSRT to represent human visual system. The final results, shown in Fig.\ref{fig:dist-on-supervised}, represent the average performance across all participating human subjects. Further detail and of the psychophysical experiments and how the experiment is conducted can be found at the supplementary. The human behavioral study was approved by the Institutional Review Board of the authors’ institution. 
Due to the high time cost of the texture synthesis process (100 optimization steps would take above 55s on a single Tesla V100 GPU, which is the time cost for a single image generation). It’s infeasible to generate a full global-relationship-distorted version of the ImageNet1k training dataset. Therefore, we use 10-step optimization results to approximate the 100-step optimization result in DiSRT.

% \newpage
\section{Results From DiSRT Benchmark}

\paragraph{Supervised Learning Models' Performance}
As a baseline, we evaluate models trained on ImageNet-1K with supervised learning signal first. From Figure~\ref{fig:dist-on-supervised}, we can see that most model perform not as well as human when trained to perform classification. This is not unreasonable since one can easily perform the classification task by just detecting the presence of local parts without worrying too much about the global structure of the image. Therefore, the models performed poorly when asked to discriminate against a global structural perturbation. However, if we compare within models trained by classification supervision, we can still see a few trends: 
\begin{figure}[htbp]
  \centering
    \centering
    \vspace{-5mm}
    \includegraphics[width=0.7\textwidth]{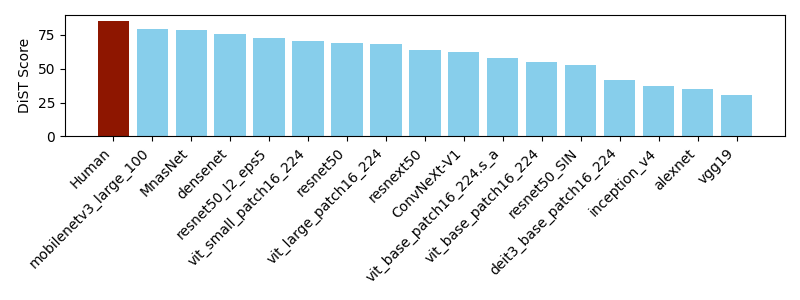}
  \caption{DiSRT score on Supervised Learning Models.}
  \label{fig:dist-on-supervised}
  \vspace{-5mm}
\end{figure}
% \vspace{-3mm}
% \begin{itemize}
    (1) ViT model are not neccessarily more sensitive than the convolutional neural networks on perceiving the global structural relationship. 
    (2) ViT small is not necessarily worse than the ViT-large or ViT-base model. 
    (3) Similarly, the lightweight mobilev3 perform better than larger convolutional networks and ViTs, suggesting that the sequeeze-excitation module may introduce global structural sensitivities, inspiring further investigation.
    (4) ConvNext-V1 is not performing too well on the benchmark. However, we will see a signficant boost in the next section when trained with MAE objective. Additionally, we also find a method (Activation Top-K) that improves the ConvNeXt-V1's DiSRT score without explicitly changing the training method.
% \end{itemize}

\begin{figure}[htbp]
  \centering
    \centering
    % \vspace{-3mm}
    \includegraphics[width=0.8\textwidth]{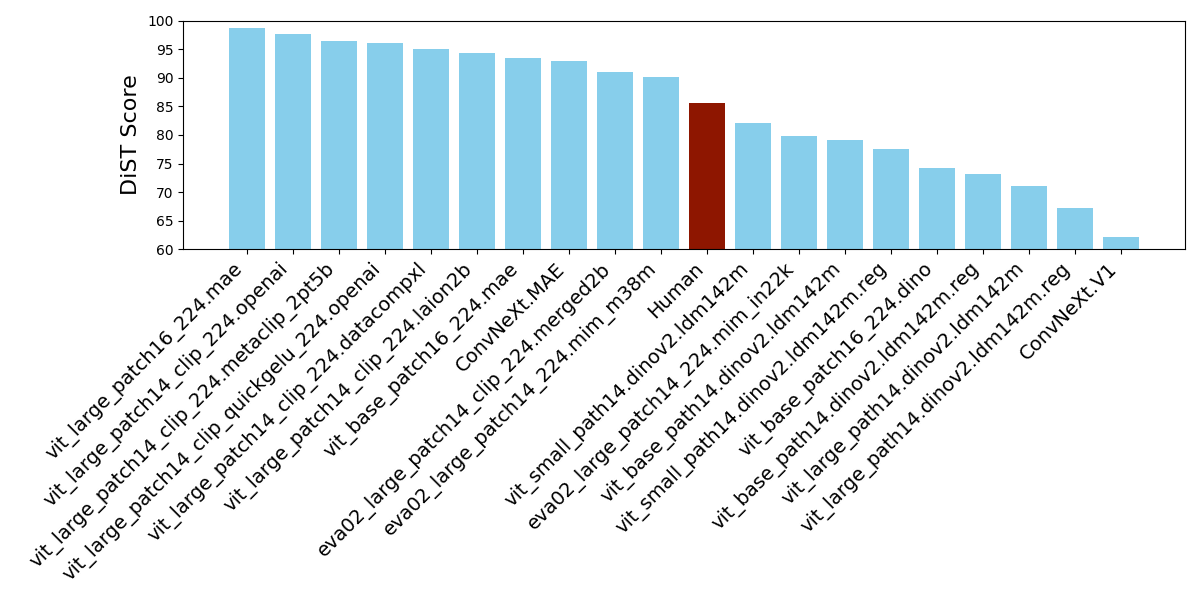}
    \vspace{-4mm}
  \caption{DiSRT score on Self-Supervised Learning Models: Various self-supervise learned ViTs. Additionally, ConvNeXt-MAE also perform significantly higher than the ConvNeXt-V1 which trained on ImageNet-1K using ConvNeXt.}
  \label{fig:dist-on-ssl}
  \vspace{-5mm}
\end{figure}

\paragraph{Power of Self-Supervise Learning}Next, we evaluate various recent models that is trained with Self-Supervised Learning method with internet-scale dataset. The results of DiSRT score is documented in Figure~\ref{fig:dist-on-ssl}. Throughout the comparison with various methods including MAE~\cite{he2022masked}, CLIP~\cite{radford2021learning}, DINOv2~\cite{oquab2023dinov2}, etc, we see that models trained with MAE and CLIP can significantly boost the performance of DiSRT score. Specifially, we can see that MAE can boost the performance of various models from under-human level to super-human level. For example, for the latest version of ConvNeXt, it incorporates MAE-style training in the training and significantly improves its stand on the DiSRT benchmark.

It is reasonable that the MAE is critical to enhance global relationship modeling, as its job is to reconstruct the missing part given the global context. 
For CLIP model on the other hand, is quite surprising. In \cite{yuksekgonul2022and}, the author suggests that the VLM models doesn't not display a coherent compositional understanding, as when they perturb the word order of the image caption, the retrieval models exhibits bag-of-words like behaviors. However, \cite{yuksekgonul2022and}  only introduce perturbation on the language side and they didn't explicitly perturb the image like ours. With our benchmark, we contribute to further understand the compositional nature of the CLIP training method. 

\begin{figure}
    \centering
\includegraphics[width=0.83\linewidth]{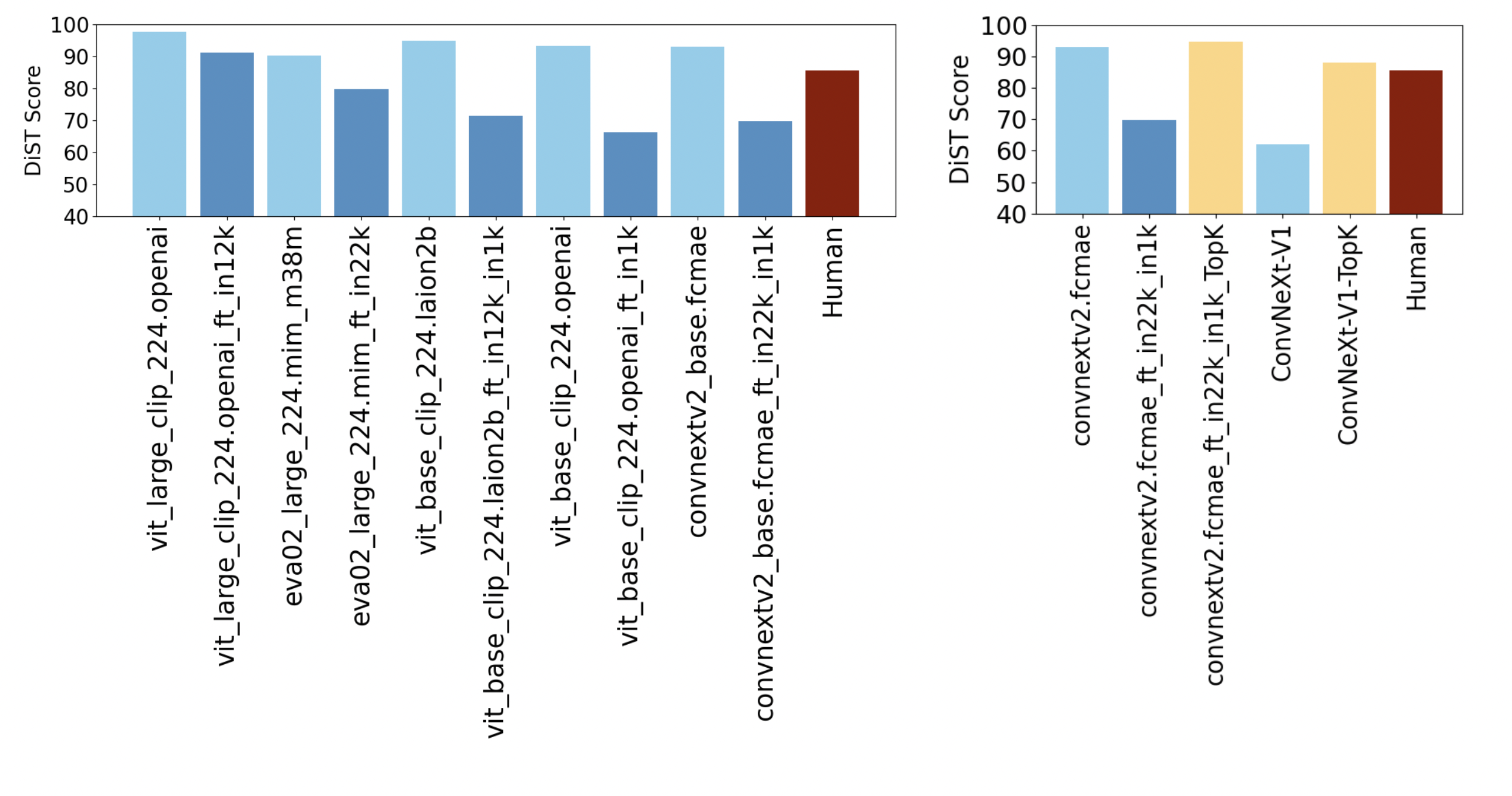}
    \vspace{-5mm}
    \caption{Curse of Classification Finetuning and Top-K Activation Sparsity. Finetuning with classification consistently decreases the DiSRT score, regardless what pretraining methods used. Adding Top-K activation constraint would revive the finetuned model to top level DiSRT rankings. }
    \vspace{-2mm}
    \label{fig:curse_of_cls_and_topk}
    \vspace{-4mm}
\end{figure}

\vspace{-2mm}
\paragraph{Curse of Classification Finetuning}
Despite the significant progress made by self-supervised learning, we see a worrying dip in the DiSRT score when model is finetuned with ImageNet-1K classification tasks regardless what pretraining method one uses (Figure~\ref{fig:curse_of_cls_and_topk} Left). The light blues are the pretrianing method with either CLIP, or MAE, whereas the darker blue right next to the light one is the classification finetuned models. A significant DiSRT performance drop when finetuning with classification task. This means that the current supervised categorical learning is fundamentally limited -- even pretraining provides good initialization, the objective function keep going into a shortcut learning case to classify image based not on the global structure, but rather relies on local textures and bag-of-words modeling.   

\vspace{-3mm}

\paragraph{Top-K Sparsity to the Rescue: Reviving Classification Finetuned Models.} In the last section, we demonstrate that Classification Finetuning would degrade the models' ability to perceive global relationships. Here, we propose a simple yet effective way to significantly improve the results. Following \cite{li2023emergence} and \cite{tang2018large} in the neuroscience literature, we apply a Top-K operation to the activation of neural networks (keep only the top activated neurons 20\% and eliminate the weakly activated neurons at the bottom 80\%, see details in the supplementary) and observe an interesting improvement (from 69.8 to 94.6 DiSRT score) of the model pre-trained with MAE but fine-tuned on ImageNet-22k and ImageNet-1k (Figure~\ref{fig:curse_of_cls_and_topk} right). Additionally, we also observe a significant increase in the model without pre-training (ConvNeXt-V1),
from 62.1 DiSRT score to 88.1 DiSRT score when applied Top-K operation to activation. We refer to the supplementary for the implementation of Top-K and the parameter setup used to bring up the performance.

% \paragraph{Giant models (600M to 1.8B)}

% \begin{figure}[htbp]
%   \centering
%     \centering
%     \includegraphics[width=0.9\textwidth]{images/model_scores_giant.png}
%   \caption{DiSRT score on Giant Models (Range from 600 M to 1.8 B).}
%   \label{fig:dist-on-giant-models}
% \end{figure}

% \newpage
\vspace{-2mm}
\section{Related Works}
\vspace{-2mm}

Deep neural networks (DNNs) have achieved remarkable success in computer vision~\citep{luo2021surfgen, he2016deep, brown2020language, ma2022robust, wu2023hallucination, tsai2021integrating, 10052697, hong2022deep, tsai2022learning, li2022prototype, tsai2022conditional}, yet understanding their internal biases remains critical~\citep{drenkow2021systematic, petch2022opening, gilpin2018explaining, liao2023assessing, kim2021neural, zhao2023evaluating, chen2025vision}. While early work suggested that DNNs primarily learn shape-based features—supported by hierarchical visualizations in CNNs~\citep{zeiler2014visualizing}—\citep{geirhos2018imagenet} showed that ImageNet-trained models exhibit a strong bias toward textural local parts. This led to style-transfer-based benchmarks~\citep{geirhos2018imagenet, geirhos2021partial}, where Vision Transformers (ViTs)\citep{dosovitskiy2020image} were found to perform better than CNNs in terms of shape bias and robustness to corruption~\citep{paul2022vision}. However, our results show that ViTs exhibit no significant improvement on our proposed DiSRT benchmark, which directly probes global spatial sensitivities. This challenges the assumption that ViTs inherently model global relationships better, and highlights the need for architectural or training innovations to improve true shape-based perception.

\vspace{-2mm}
\section{Discussion \& Conclusion}
\vspace{-2mm}

This work provides a systematic investigation into the emergence of Gestalt perception in deep vision models, particularly those trained via self-supervised learning. Through both qualitative analysis using classic visual illusions and a novel quantitative benchmark—DiSRT—we demonstrate that models trained with masked autoencoding (MAE) or CLIP objectives develop strong sensitivity to global visual structure, enabling perceptual behaviors aligned with Gestalt principles such as closure, proximity, continuity, and figure–ground organization.

Our findings reveal three central insights:
(1) Gestalt-like perception is not exclusive to Vision Transformers—convolutional models trained with MAE also exhibit similar organization, indicating the importance of training objectives over architectural priors;
(2) Supervised finetuning degrades global dependency sensitivity, suggesting that standard classification tasks may suppress perceptual grouping abilities;
(3) Top-K activation sparsity can revive these sensitivities, offering a simple, biologically inspired mechanism to restore structured perception in post-finetuned models.

While our study establishes a link between global structure sensitivity and Gestalt-like behaviors, it remains correlational rather than causal. Our analysis relies on PCA-based probes and handcrafted stimuli, which—though insightful—do not capture the full complexity of natural perception. Additionally, DiSRT evaluates one specific form of global dependency (spatial configuration), and may not fully represent the diversity of perceptual grouping mechanisms observed in humans.

In summary, by connecting modern deep learning methods to classical theories of perception—such as Marr’s hierarchical vision model and the Gestalt school—we bridge a gap between computational and cognitive perspectives. Our benchmark, DiSRT, provides a scalable and architecture-agnostic tool for probing global perception in neural networks.

\bibliographystyle{unsrt}

\bibliography{ref}

\begin{thebibliography}{10}

\bibitem{wagemans2012centuryI}
Johan Wagemans, James~H Elder, Michael Kubovy, Stephen~E Palmer, Mary~A Peterson, Manish Singh, and Rüdiger von~der Heydt.
\newblock A century of gestalt psychology in visual perception: I. perceptual grouping and figure–ground organization.
\newblock {\em Psychological Bulletin}, 138(6):1172--1217, 2012.

\bibitem{wagemans2012centuryII}
Johan Wagemans, Jacob Feldman, Sergei Gepshtein, Ruth Kimchi, James~R Pomerantz, Peter~A van~der Helm, and Cees van Leeuwen.
\newblock A century of gestalt psychology in visual perception: Ii. conceptual and theoretical foundations.
\newblock {\em Psychological Bulletin}, 138(6):1218--1252, 2012.

\bibitem{wertheimer1923}
Max Wertheimer.
\newblock Untersuchungen zur lehre von der gestalt, ii.
\newblock {\em Psychologische Forschung}, 4:301--350, 1923.

\bibitem{kanizsa1976subjective}
Gaetano Kanizsa.
\newblock Subjective contours.
\newblock {\em Scientific American}, 234(4):48--52, 1976.

\bibitem{he2022masked}
Kaiming He, Xinlei Chen, Saining Xie, Yanghao Li, Piotr Doll{\'a}r, and Ross Girshick.
\newblock Masked autoencoders are scalable vision learners.
\newblock In {\em Proceedings of the IEEE/CVF conference on computer vision and pattern recognition}, pages 16000--16009, 2022.

\bibitem{arnheim1972art}
Rudolf Arnheim.
\newblock {\em Art and visual perception: A psychology of the creative eye}.
\newblock Univ of California Press, 1972.

\bibitem{kanizsa1976convexity}
Gaetano Kanizsa, Rudolf ARNHEIM, Mary HENLE, and Walter GERBINO.
\newblock Convexity and symmetry in figure-ground organization.
\newblock In {\em Vision and artifact}, 1976.

\bibitem{peterson2008inhibitory}
Mary~A Peterson and Elizabeth Salvagio.
\newblock Inhibitory competition in figure-ground perception: Context and convexity.
\newblock {\em Journal of Vision}, 8(16):4--4, 2008.

\bibitem{bertamini2006owns}
Marco Bertamini.
\newblock Who owns the contour of a visual hole?
\newblock {\em Perception}, 35(7):883--894, 2006.

\bibitem{liu2022convnet}
Zhuang Liu, Hanzi Mao, Chao-Yuan Wu, Christoph Feichtenhofer, Trevor Darrell, and Saining Xie.
\newblock A convnet for the 2020s.
\newblock In {\em Proceedings of the IEEE/CVF conference on computer vision and pattern recognition}, pages 11976--11986, 2022.

\bibitem{woo2023convnext}
Sanghyun Woo, Shoubhik Debnath, Ronghang Hu, Xinlei Chen, Zhuang Liu, In~So Kweon, and Saining Xie.
\newblock Convnext v2: Co-designing and scaling convnets with masked autoencoders.
\newblock In {\em Proceedings of the IEEE/CVF conference on computer vision and pattern recognition}, pages 16133--16142, 2023.

\bibitem{gatys2015texture}
Leon Gatys, Alexander~S Ecker, and Matthias Bethge.
\newblock Texture synthesis using convolutional neural networks.
\newblock {\em Advances in neural information processing systems}, 28, 2015.

\bibitem{radford2021learning}
Alec Radford, Jong~Wook Kim, Chris Hallacy, Aditya Ramesh, Gabriel Goh, Sandhini Agarwal, Girish Sastry, Amanda Askell, Pamela Mishkin, Jack Clark, et~al.
\newblock Learning transferable visual models from natural language supervision.
\newblock In {\em International conference on machine learning}, pages 8748--8763. PmLR, 2021.

\bibitem{oquab2023dinov2}
Maxime Oquab, Timoth{\'e}e Darcet, Th{\'e}o Moutakanni, Huy Vo, Marc Szafraniec, Vasil Khalidov, Pierre Fernandez, Daniel Haziza, Francisco Massa, Alaaeldin El-Nouby, et~al.
\newblock Dinov2: Learning robust visual features without supervision.
\newblock {\em arXiv preprint arXiv:2304.07193}, 2023.

\bibitem{yuksekgonul2022and}
Mert Yuksekgonul, Federico Bianchi, Pratyusha Kalluri, Dan Jurafsky, and James Zou.
\newblock When and why vision-language models behave like bags-of-words, and what to do about it?
\newblock {\em arXiv preprint arXiv:2210.01936}, 2022.

\bibitem{li2023emergence}
Tianqin Li, Ziqi Wen, Yangfan Li, and Tai~Sing Lee.
\newblock Emergence of shape bias in convolutional neural networks through activation sparsity.
\newblock {\em Advances in Neural Information Processing Systems}, 36:71755--71766, 2023.

\bibitem{tang2018large}
Shiming Tang, Yimeng Zhang, Zhihao Li, Ming Li, Fang Liu, Hongfei Jiang, and Tai~Sing Lee.
\newblock Large-scale two-photon imaging revealed super-sparse population codes in the v1 superficial layer of awake monkeys.
\newblock {\em Elife}, 7:e33370, 2018.

\bibitem{luo2021surfgen}
Andrew Luo, Tianqin Li, Wen-Hao Zhang, and Tai~Sing Lee.
\newblock Surfgen: Adversarial 3d shape synthesis with explicit surface discriminators.
\newblock In {\em Proceedings of the IEEE/CVF International Conference on Computer Vision}, pages 16238--16248, 2021.

\bibitem{he2016deep}
Kaiming He, Xiangyu Zhang, Shaoqing Ren, and Jian Sun.
\newblock Deep residual learning for image recognition.
\newblock In {\em Proceedings of the IEEE conference on computer vision and pattern recognition}, pages 770--778, 2016.

\bibitem{brown2020language}
Tom Brown, Benjamin Mann, Nick Ryder, Melanie Subbiah, Jared~D Kaplan, Prafulla Dhariwal, Arvind Neelakantan, Pranav Shyam, Girish Sastry, Amanda Askell, et~al.
\newblock Language models are few-shot learners.
\newblock {\em Advances in neural information processing systems}, 33:1877--1901, 2020.

\bibitem{ma2022robust}
Wufei Ma, Angtian Wang, Alan Yuille, and Adam Kortylewski.
\newblock Robust category-level 6d pose estimation with coarse-to-fine rendering of neural features.
\newblock In {\em European Conference on Computer Vision}, pages 492--508. Springer, 2022.

\bibitem{wu2023hallucination}
Jing Wu, Jennifer Hobbs, and Naira Hovakimyan.
\newblock Hallucination improves the performance of unsupervised visual representation learning.
\newblock In {\em Proceedings of the IEEE/CVF International Conference on Computer Vision}, pages 16132--16143, 2023.

\bibitem{tsai2021integrating}
Yao-Hung~Hubert Tsai, Tianqin Li, Weixin Liu, Peiyuan Liao, Ruslan Salakhutdinov, and Louis-Philippe Morency.
\newblock Integrating auxiliary information in self-supervised learning.
\newblock {\em arXiv preprint arXiv:2106.02869}, 2021.

\bibitem{10052697}
Meng Xiao, Ziyue Qiao, Yanjie Fu, Hao Dong, Yi~Du, Pengyang Wang, Hui Xiong, and Yuanchun Zhou.
\newblock Hierarchical interdisciplinary topic detection model for research proposal classification.
\newblock {\em IEEE Transactions on Knowledge and Data Engineering}, 35(9):9685--9699, 2023.

\bibitem{hong2022deep}
Jiazhen Hong, Foroogh Shamsi, and Laleh Najafizadeh.
\newblock A deep learning framework based on dynamic channel selection for early classification of left and right hand motor imagery tasks.
\newblock In {\em 2022 44th Annual International Conference of the IEEE Engineering in Medicine \& Biology Society (EMBC)}, pages 3550--3553. IEEE, 2022.

\bibitem{tsai2022learning}
Yao-Hung~Hubert Tsai, Tianqin Li, Weixin Liu, Peiyuan Liao, Ruslan Salakhutdinov, and Louis-Philippe Morency.
\newblock Learning weakly-supervised contrastive representations.
\newblock {\em arXiv preprint arXiv:2202.06670}, 2022.

\bibitem{li2022prototype}
Tianqin Li, Zijie Li, Harold Rockwell, Amir Farimani, and Tai~Sing Lee.
\newblock Prototype memory and attention mechanisms for few shot image generation.
\newblock In {\em Proceedings of the eleventh international conference on learning representations}, volume~18, 2022.

\bibitem{tsai2022conditional}
Yao-Hung~Hubert Tsai, Tianqin Li, Martin~Q Ma, Han Zhao, Kun Zhang, Louis-Philippe Morency, and Ruslan Salakhutdinov.
\newblock Conditional contrastive learning with kernel.
\newblock {\em arXiv preprint arXiv:2202.05458}, 2022.

\bibitem{drenkow2021systematic}
Nathan Drenkow, Numair Sani, Ilya Shpitser, and Mathias Unberath.
\newblock A systematic review of robustness in deep learning for computer vision: Mind the gap?
\newblock {\em arXiv preprint arXiv:2112.00639}, 2021.

\bibitem{petch2022opening}
Jeremy Petch, Shuang Di, and Walter Nelson.
\newblock Opening the black box: the promise and limitations of explainable machine learning in cardiology.
\newblock {\em Canadian Journal of Cardiology}, 38(2):204--213, 2022.

\bibitem{gilpin2018explaining}
Leilani~H Gilpin, David Bau, Ben~Z Yuan, Ayesha Bajwa, Michael Specter, and Lalana Kagal.
\newblock Explaining explanations: An overview of interpretability of machine learning.
\newblock In {\em 2018 IEEE 5th International Conference on data science and advanced analytics (DSAA)}, pages 80--89. IEEE, 2018.

\bibitem{liao2023assessing}
Danqi Liao, Chen Liu, Alexander Tong, Guillaume Huguet, Guy Wolf, Maximilian Nickel, Ian Adelstein, and Smita Krishnaswamy.
\newblock Assessing neural network representations during training using data diffusion spectra.
\newblock {\em ICML workshop on TAG-ML}, 2023.

\bibitem{kim2021neural}
Been Kim, Emily Reif, Martin Wattenberg, Samy Bengio, and Michael~C Mozer.
\newblock Neural networks trained on natural scenes exhibit gestalt closure.
\newblock {\em Computational Brain \& Behavior}, 4(3):251--263, 2021.

\bibitem{zhao2023evaluating}
Zhuokai Zhao, Takumi Matsuzawa, William Irvine, Michael Maire, and Gordon~L Kindlmann.
\newblock Evaluating machine learning models with nero: Non-equivariance revealed on orbits.
\newblock {\em arXiv preprint arXiv:2305.19889}, 2023.

\bibitem{chen2025vision}
Yinjie Chen, Zipeng Yan, Chong Zhou, Bo~Dai, and Andrew~F Luo.
\newblock Vision transformers with self-distilled registers.
\newblock {\em arXiv preprint arXiv:2505.21501}, 2025.

\bibitem{zeiler2014visualizing}
Matthew~D Zeiler and Rob Fergus.
\newblock Visualizing and understanding convolutional networks.
\newblock In {\em Computer Vision--ECCV 2014: 13th European Conference, Zurich, Switzerland, September 6-12, 2014, Proceedings, Part I 13}, pages 818--833. Springer, 2014.

\bibitem{geirhos2018imagenet}
Robert Geirhos, Patricia Rubisch, Claudio Michaelis, Matthias Bethge, Felix~A Wichmann, and Wieland Brendel.
\newblock Imagenet-trained cnns are biased towards texture; increasing shape bias improves accuracy and robustness.
\newblock {\em arXiv preprint arXiv:1811.12231}, 2018.

\bibitem{geirhos2021partial}
Robert Geirhos, Kantharaju Narayanappa, Benjamin Mitzkus, Tizian Thieringer, Matthias Bethge, Felix~A Wichmann, and Wieland Brendel.
\newblock Partial success in closing the gap between human and machine vision.
\newblock {\em Advances in Neural Information Processing Systems}, 34:23885--23899, 2021.

\bibitem{dosovitskiy2020image}
Alexey Dosovitskiy, Lucas Beyer, Alexander Kolesnikov, Dirk Weissenborn, Xiaohua Zhai, Thomas Unterthiner, Mostafa Dehghani, Matthias Minderer, Georg Heigold, Sylvain Gelly, et~al.
\newblock An image is worth 16x16 words: Transformers for image recognition at scale.
\newblock {\em arXiv preprint arXiv:2010.11929}, 2020.

\bibitem{paul2022vision}
Sayak Paul and Pin-Yu Chen.
\newblock Vision transformers are robust learners.
\newblock In {\em Proceedings of the AAAI conference on Artificial Intelligence}, volume~36, pages 2071--2081, 2022.

\bibitem{deng2009imagenet}
Jia Deng, Wei Dong, Richard Socher, Li-Jia Li, Kai Li, and Li~Fei-Fei.
\newblock Imagenet: A large-scale hierarchical image database.
\newblock In {\em 2009 IEEE conference on computer vision and pattern recognition}, pages 248--255. Ieee, 2009.

\bibitem{touvron2022deit}
Hugo Touvron, Matthieu Cord, and Herv{\'e} J{\'e}gou.
\newblock Deit iii: Revenge of the vit.
\newblock In {\em European conference on computer vision}, pages 516--533. Springer, 2022.

\bibitem{laion_clip_vit_h_14_2022}
Romain Beaumont.
\newblock Clip-vit-h-14-laion2b-s32b-b79k.
\newblock \url{https://huggingface.co/laion/CLIP-ViT-H-14-laion2B-s32B-b79K}, 2022.
\newblock Model trained using OpenCLIP on the LAION-2B dataset.

\end{thebibliography}
% \title{Appendix}
\newpage
\begin{center}
    \LARGE \textbf{Supplementary Materials}
\end{center}

\section*{A ~~ Intermediate Representation Analysis}

To better understand why certain methods provide superior global relationship sensitivities, we compute the DiSRT score based on the intermediate activation of the model. If one consider the ViT utilizes attention to integrate global information, a natural way to gain insight of the internal activation of each layer is to perform DiSRT analysis for the activation of each attention head right after it integrates global context. We will first start with models trained with supervised learning. 

\begin{figure}[htbp]
    \centering
    \includegraphics[width=\linewidth]{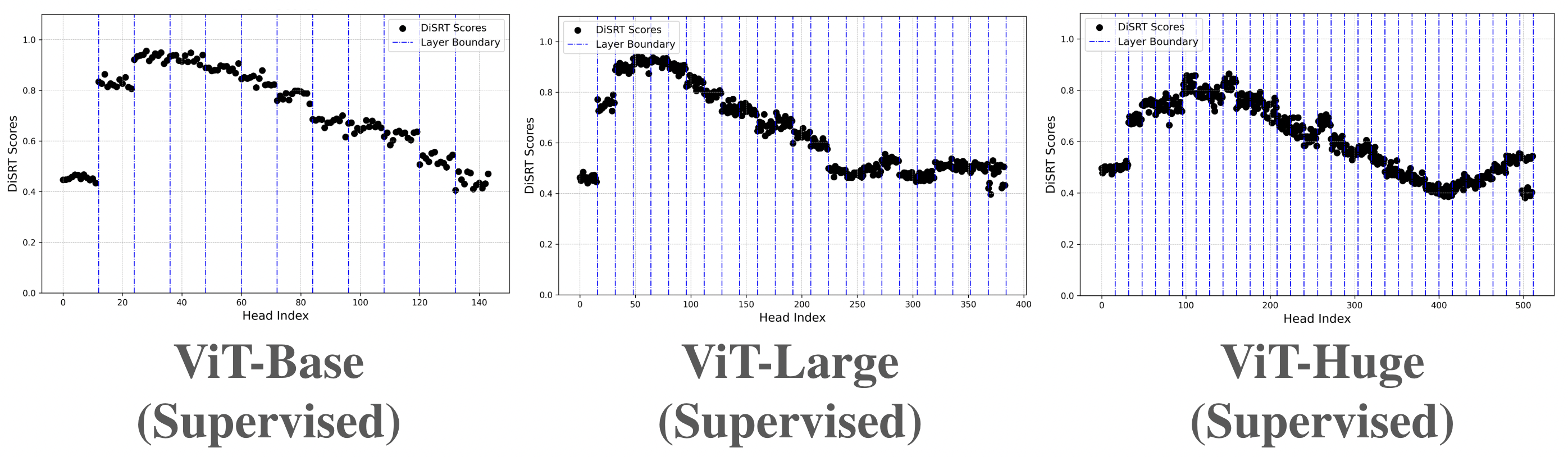}
    \caption{DiSRT score for the intermediate layer of supervised learning trained ViTs. The ViT-B/L/H are from the original proposed ViT paper~\cite{dosovitskiy2020image} trained on ImageNet-21K dataset~\cite{deng2009imagenet}. Across the ViT architecture, attention head is indexed based on its layer number and its original index inside the layer. Layer boundary is highlighted using blue dash lines, for example the first 12 head in the ViT-Base belongs to the first layer, therefore there is a blue dash line at the Head Index 12. From the plot we can see a trend of first increased then decreased global structural sensitivities going from the early layer to the deeper layers.}
    \label{fig:supervised_internal}
\end{figure}

\subsection*{A.1 ~~ Supervised Learning}
\begin{figure}
    \centering
    \includegraphics[width=\linewidth]{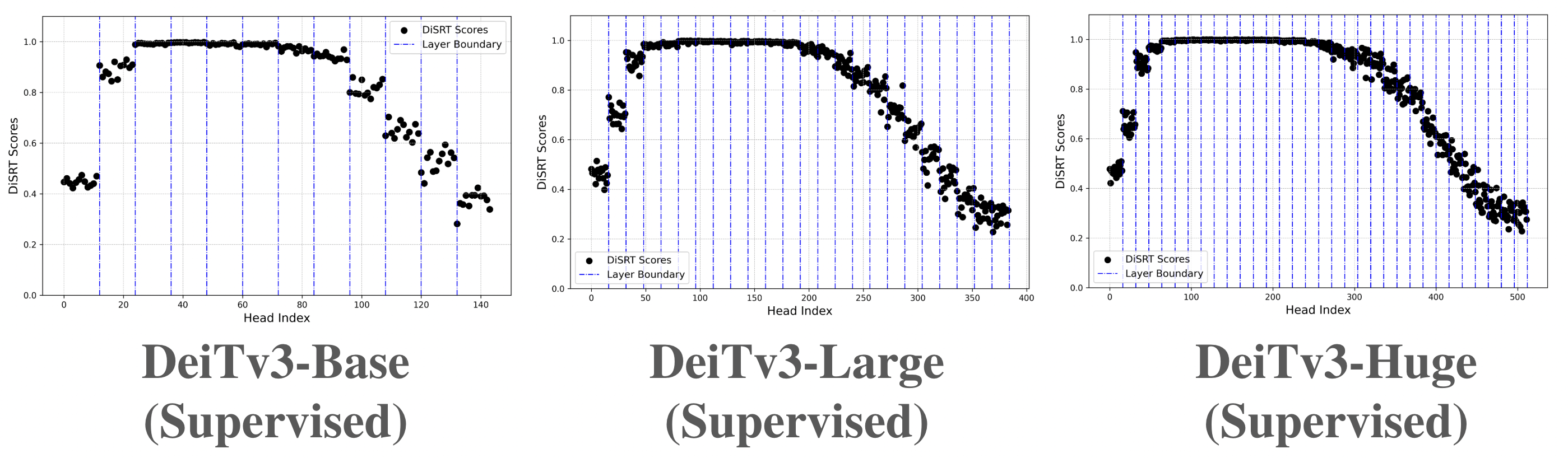}
    \caption{DiSRT score for another supervised learning variant, DeiT v3~\cite{touvron2022deit}. We can observe a similar trend where the early and the middle layer layers learned to perceive global structure but gradually lose this information during the bottom-up process.}
    \label{fig:supervised_internal_deit}
\end{figure}

In the ViT architecture, we index each attention head sequentially by concatenating all heads across layers, where the index reflects both the head’s position within its layer and the layer’s depth. In Figure~\ref{fig:supervised_internal}, we plot the DiSRT score evaluated for each attention head. To indicate layer boundaries, we insert vertical blue dashed lines—for instance, in ViT-Base, the first 12 heads correspond to the first transformer layer, hence a dashed line is placed at Head Index 12. From Figure~\ref{fig:supervised_internal}, we can easily spot the trend of first increase and then decrease of the global structural sensitivities. It is surprising since from the results itself, it actually knows quite a bit of the global structural sensitivities in the early layer (almost 100\%), however such capability gradually decreases as the layer goes deeper. Such phenomenon can be observed regardless of the model size (ViT-B/ViT-L/ViT-H). Combining with the results of the Figure 7 in main text where we showed that supervised learned model is bad at the global sensitivity test, it ponders the question of why the information was once captured but later on was "given-up" in the bottom-up process. One potential explanation is that the supervised learning objective didn't consider it is very important to keep such information in the final classification task. This raises further concerns about the way machine understand images as they are not relying on the global structure of the objects to perform classification, but rather decreasing its sensitivities of them as layers become more semantic. We can observe similar trend of another more recent supervised learning method -- DeiT v3~\cite{touvron2022deit} where different augmentation and training hyperparameters are optimized (See~\ref{fig:supervised_internal_deit}).

\subsection*{A.2 ~~ Generative Pretraining (MAE) Maintains the Global Structural Awareness across Layers}
\begin{figure}[htbp]
    \centering
    \includegraphics[width=\linewidth]{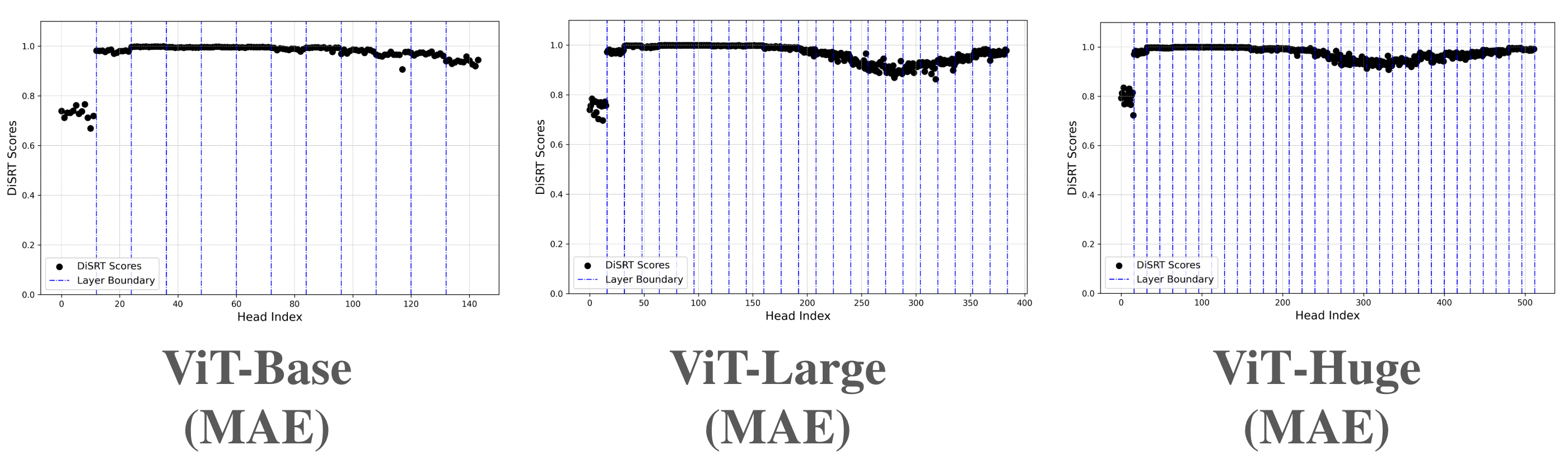}
    \caption{DiSRT score for different architectures trained on MAE~\cite{he2022masked}. We can clearly see a difference in MAE as it almost steadily maintain the sensitivities to global structure perturbation. We hypothesize that it is such behavior during the bottom-up phase make it rank the top of the DiSRT benchmarks. }
    \label{fig:MAE_intermediate}
\end{figure}

Next, we test the intermediate layers of MAEs. Surprisingly, we see very little DiSRT degradation across the bottom-up layers in models trained with MAE as shown in Figure~\ref{fig:MAE_intermediate}. Comparing models trained with supervised classification loss (Figure~\ref{fig:supervised_internal_deit} and Figure~\ref{fig:supervised_internal}), MAE objectives force the intermediate layers to maintain the global structure information, which explains its superior performance when its final representation is tested. 

\subsection*{A.2 ~~ Impact of CLIP Pretraining}
\begin{figure}[htbp]
    \centering
    \includegraphics[width=\linewidth]{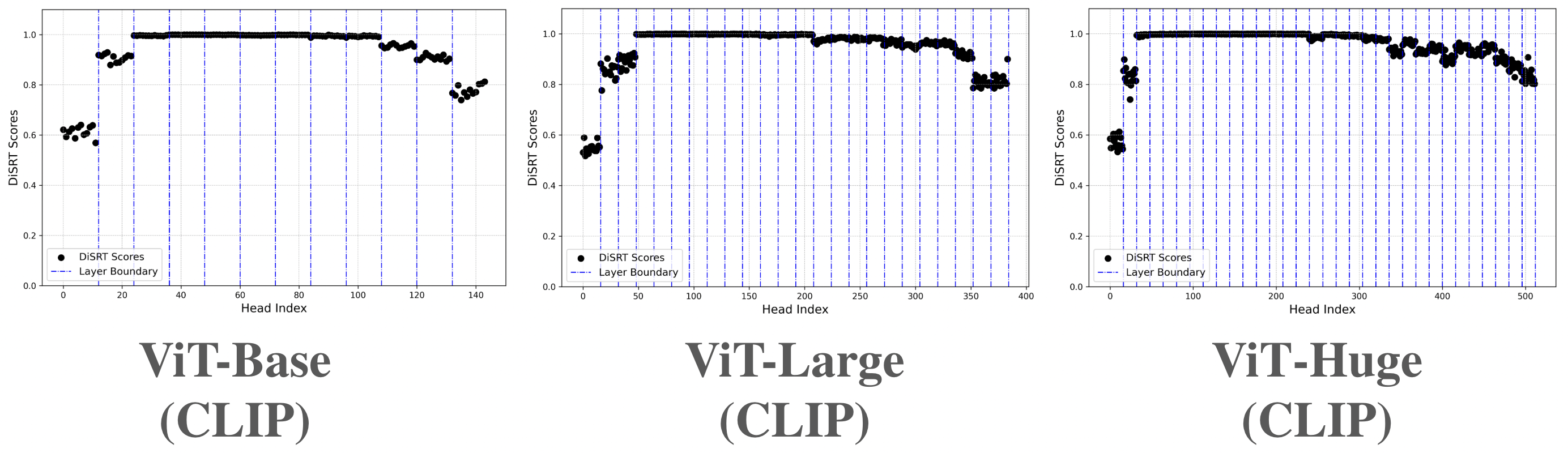}
    \caption{DiSRT score for different architectures trained on CLIP with LAION-2B dataset~\cite{laion_clip_vit_h_14_2022}. It holds up the global structural sensitivities much better than the supervised learning one but still lose the capability at the last few layers.}
    \label{fig:clip-intermediate}
\end{figure}

Since CLIP is also an important method that yields good performance, we investigate how does the internal DiSRT score evolve across different layers in CLIP trained ViTs. We utilizes models that are trained on LAION-2B dataset with CLIP method~\cite{laion_clip_vit_h_14_2022} and show the results in Figure~\ref{fig:clip-intermediate}. Although CLIP performs relatively well comparing to the supervised learning benchmark on DiSRT, it still lose a little bit of global structural sensitivities when going into deeper layers. Given that the models evaluated upon are trained on LAION-2B dataset while MAE in Figure~\ref{fig:MAE_intermediate} are only trained on ImageNet-scale, it suggests that MAE is much more effective in equipping models with global structural awareness across the layers. 

\subsection*{A.3 ~~ DiSRT on DINOv2 Intermediate Representations}
\begin{figure}[htbp]
    \centering
    \includegraphics[width=\linewidth]{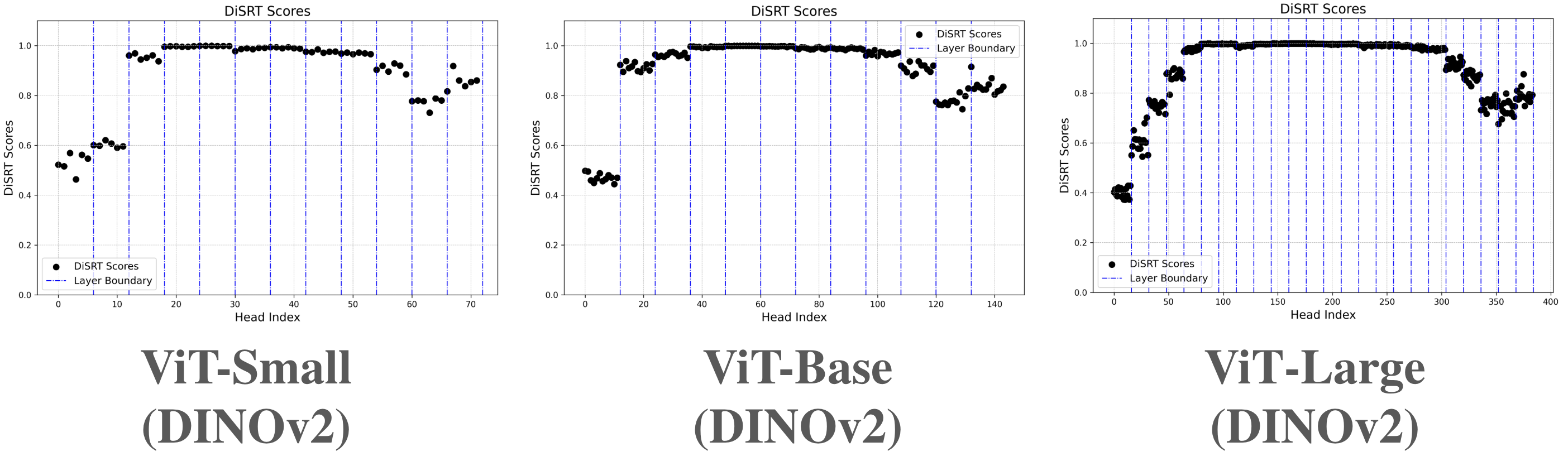}
    \caption{DiSRT score on DINOv2 architectures.}
    \label{fig:dinov2-intermediate}
\end{figure}

DINOv2~\cite{oquab2023dinov2} is an state-of-the-art pure image based pretraining method. In this section, we want to understand how well DINOv2 would preserve the global structures across the layers. We can see from the Figure~\ref{fig:dinov2-intermediate} that the ability to capture global structure thrives in the intermediate layers, but at the last few layers of DINOv2, it degrades a little bit. This could explain what we observe in the main text Figure 8 where DINOv2 perform significantly worse than MAE and CLIP trained models despite trained with 142 Million curated dataset. Since DINOv2 has two component -- a contrastive learning objective and a mask image modeling component, one further investigation direction is to see why combing mask image modeling with contrastive learning hinders the performance of global structural awareness while apply them seperately (MAE or CLIP) would produce superior results on DiSRT.

\subsection*{A.4 ~~ Impact of Supervised Finetuning in the Intermediate Layers}
\begin{figure}[htbp]
    \centering
    \includegraphics[width=0.8\linewidth]{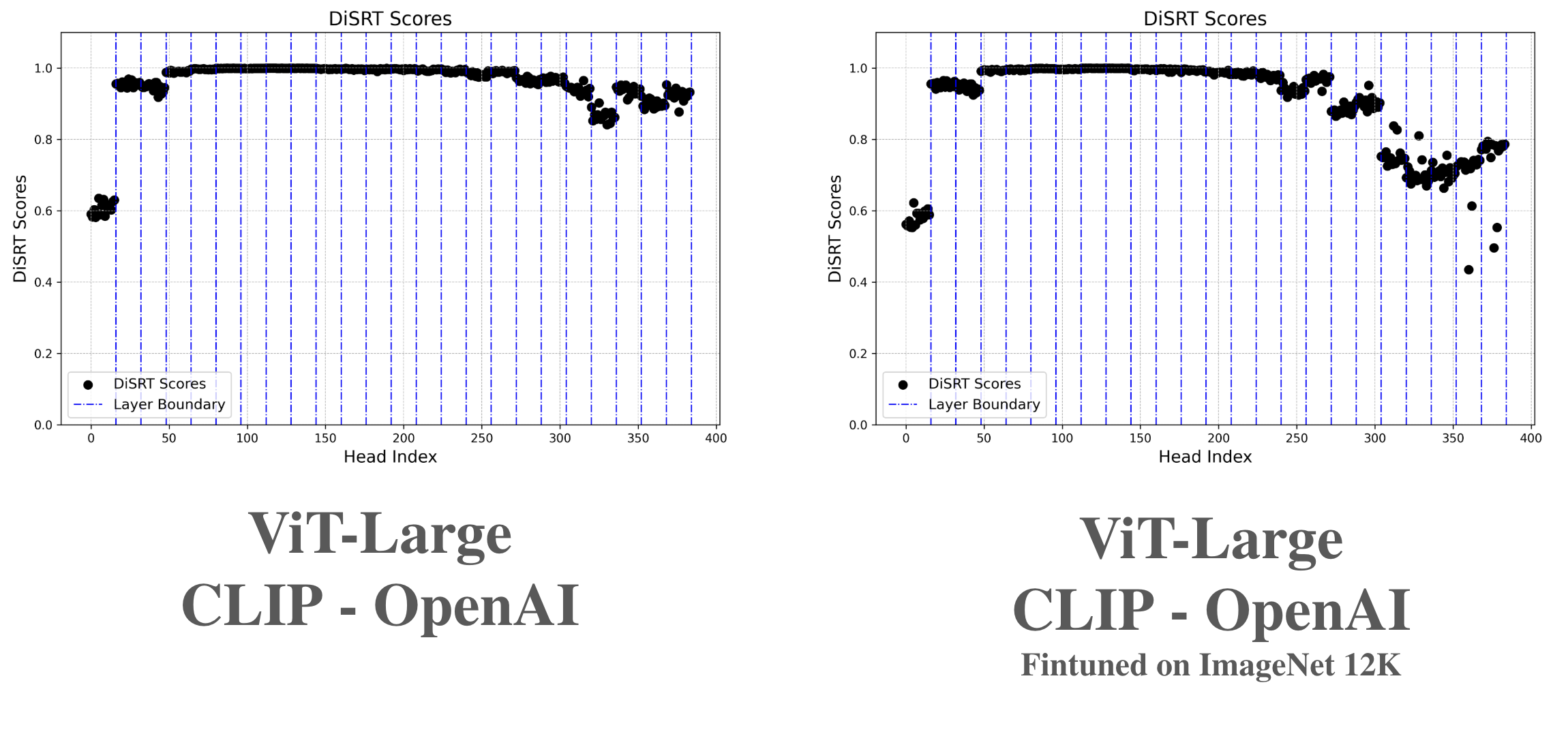}
    \caption{DiSRT score comparison before and after supervised finetuning on ViT-Large CLIP trained models. The last few layers are impacted. }
    \vspace{-3mm}\label{fig:impact_finetunining_before_after}
\end{figure}
As illustrated in Figure~\ref{fig:impact_finetunining_before_after}, supervised finetuning for classification substantially diminishes the model's ability to encode global structural information, particularly in the later layers. This degradation suggests that the discriminative signals introduced during finetuning may overwrite the global structure-sensitive representations acquired during pretraining. Notably, the DiSRT scores decline in the upper layers, indicating a loss of long-range dependency processing, which is essential for Gestalt-aligned perception. This raises the question of what would be a better way to maintain the global structure awareness in the last few layers during finetuning. This analysis also points out that the middle layer might not be affected during the supervised fine-tuning. 

\subsection*{A.5 ~~ Impact of Top-K sparsity in convolutional neural networks}

\begin{figure}[htbp]
    \centering
    \includegraphics[width=\linewidth]{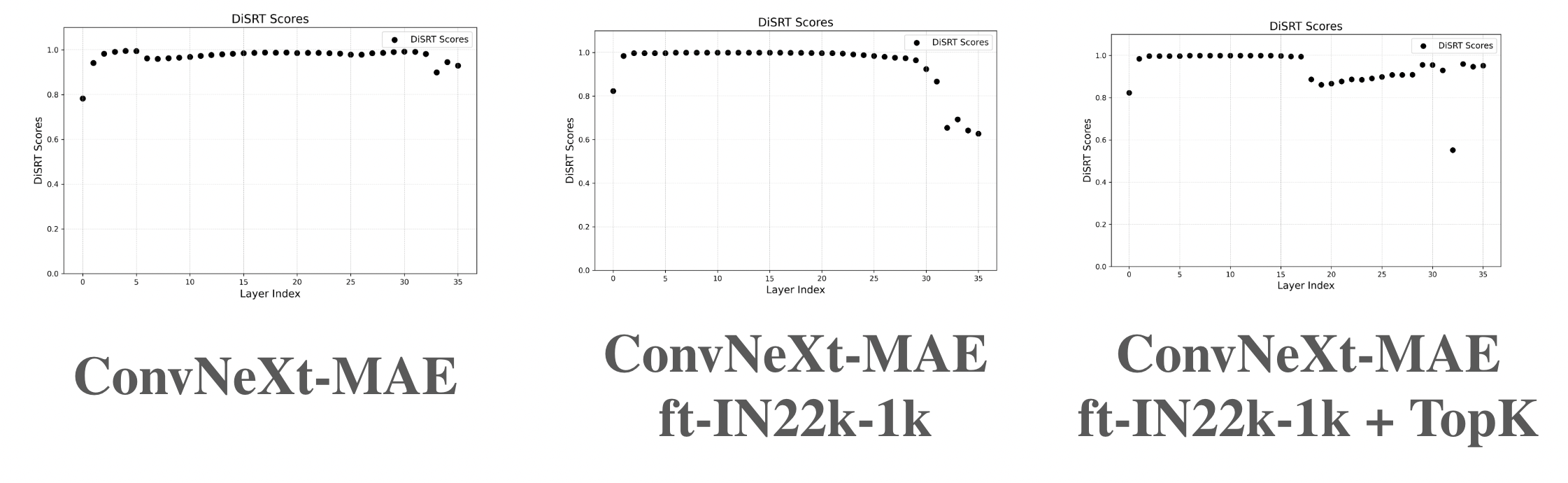}
    \caption{DiSRT score on ConvNeXt-MAE, the ImageNet finetuned variant and Top-K improved version. The DiSRT score in the last few layers drops significantly in supervised finetuning models but with Top-K applied, the final DiSRT score can be improved  significantly.}
    \label{fig:disrt_score_topk}
\end{figure}

The DiSRT scores for ConvNeXt-MAE, its ImageNet-finetuned variant, and the Top-K enhanced version reveal a notable trend: supervised finetuning leads to a significant drop in DiSRT scores in the final layers, indicating a loss of global structure sensitivity. However, when Top-K sparsification is applied, this degradation is effectively mitigated, and the final DiSRT score improves substantially, suggesting that Top-K helps preserve or recover global dependency representation. See results in Figure~\ref{fig:disrt_score_topk}.

\section*{B ~~ Methodology for Visualizing ViT-MAE Activations (Figure 1 and 2 in Main Text)}

We extract activation maps from ViT-B model trained by MAE objective on ImageNet~\cite{he2022masked}. Each activation tensor $A \in \mathbb{R}^{C \times H \times W}$ is flattened across spatial dimensions, standardized, and projected onto the first 3 principal components. These projections are then visualized using RGB mapping.
The PCA basis is shared across all visualizations to ensure interpretability.

\section*{C ~~ Interpreting PCA Projection as Figure–Ground Separation (Figure 3 in Main Text)}

\begin{figure}[htbp]
    \centering
    \includegraphics[width=\linewidth]{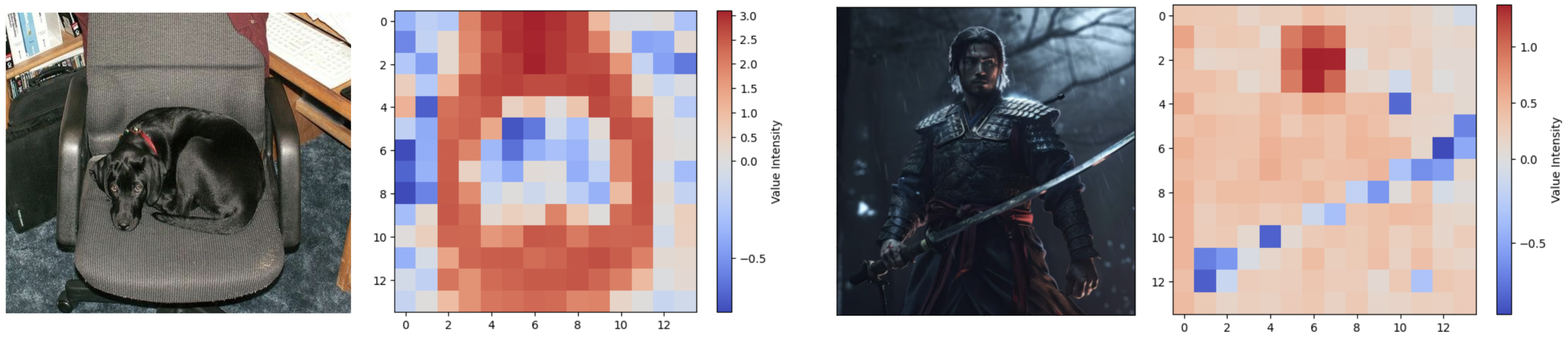}
    \caption{PCA projection value indicating figure and ground assignment. Front is represented by negative value (blue) while background is represented by positive value (orange)}
    \label{fig:front-back-classification}
\end{figure}

We validate that negative PCA projection values correspond to figure regions by analyzing activations on natural images. In Figure~\ref{fig:front-back-classification}, we demonstrate on natural image that the positive / negative of PCA projection value projected onto the first PC can be used as an proxy for figure ground assignment. The value negative indicates the figure or front whereas the positive value indicates the background assignment. 

% \section{Layer Wise Analysis on Figure Ground Segmentation}

% Here we demonstrate a detailed analysis on using PASCAL VOC subset to label the figure ground distribution and compute a quantitative measure about the distribution discrepancy for each attention head. 

% \begin{figure}
%     \centering
%     \includegraphics[width=0.5\linewidth]{}
%     \caption{Caption}
%     \label{fig:enter-label}
% \end{figure}

\section*{D ~~ Implementation Details for Top-K Activation Sparsity}

We define Top-K activation sparsity as:
\[
A_{\text{TopK}} = A \cdot \mathbb{I}[A \geq \text{threshold}_K]
\]
where $\text{threshold}_K$ is the $K^{\text{th}}$ top K percentile of $A$ across a single convolution channel. For example, support we have a specific convolutional channel $A_c = R^{H \times W}$ ($c \in \{1, 2, ...\})$, $\text{threshold}_K$ = \texttt{Rank(Flatten($A_c$))[K]}

\subsection*{D.1 ~~ Implementation (PyTorch)}
\begin{verbatim}
class TopKLayerImplementation(nn.Module):
    def __init__(self, topk=0.1):
        super(TopKLayerImplementation, self).__init__()
        self.topk=topk

    def sparse_hw(self, x, topk, device='cuda'):
        n, c, h, w = x.shape
        if topk == 1:
            return x
        x_reshape = x.view(n, c, h * w)
        topk_keep_num = int(max(1, topk * h * w))
        _, index = torch.topk(x_reshape.abs(), topk_keep_num, dim=2)
        mask = torch.zeros_like(x_reshape).scatter_(2, index, 1).to(device)
        sparse_x = mask * x_reshape
        return sparse_x.view(n, c, h, w) 
       
    def forward(self, x):
        return self.sparse_hw(x, self.topk)
\end{verbatim}

\subsection*{D.2 ~ TopK configuration}
To reproduce the Figure~2 in the main manuscript, one can employ the following TopK configuration: (1) For \texttt{convnextv2.fcmae\_ft\_in22k\_in1k\_TopK}, we apply topK to the layers in the second stage of the ConvNeXt network. The block number within the second stage and the correponding topK sparsity can be found here: \texttt{\{'1': 0.2, '2': 0.25, '5': 0.25, '6': 0.25, '23': 0.3, '24': 0.25\}}, where it follows the format of \texttt{\{block\_number : sparsity\_enforced\}}. For reviving the ConvNext-V1 (i.e. the ConvNeXt-V1-TopK in main manuscript Figure 9 Right), we find that a simple uniform application of sparsity of 0.2 to all the blocks in the stage 2 would be sufficient. 

\section*{E ~~ Giant models (600M to 1.8B)}
We also include models with very large number of parameters for reference. We can observe that it's not nessarily that larger models consistently yield superior results. The training method plays a deciding role in DiSRT measurements. For example, the best model is ViT-Huge with only 600 M parameters trained with MAE, while it still performs better than the  \texttt{vit\_gigantic\_patch14\_clip\_224} which is an 1.8 B parameters model with CLIP training method. 
\begin{figure}[htbp]
  \centering
    \centering
    \includegraphics[width=0.9\textwidth]{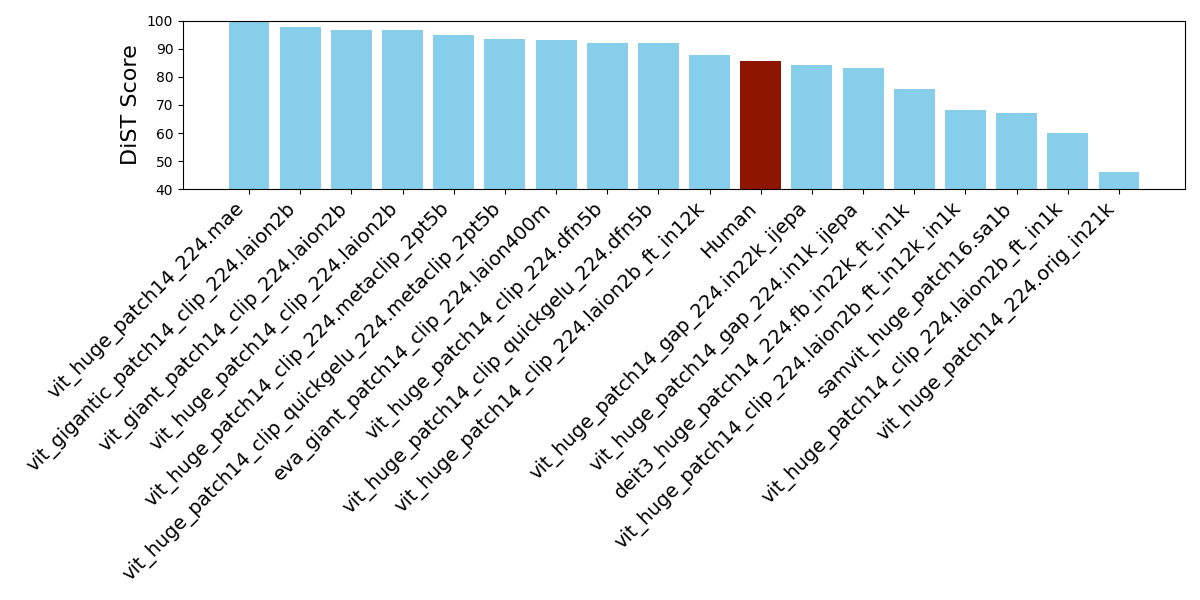}
  \caption{DiSRT score on Giant Models (Range from 600 M to 1.8 B).}
  \label{fig:dist-on-giant-models}
  \vspace{-5mm}
\end{figure}

\section*{F ~~ Broader Impacts and Safeguards}

\subsection*{F.1 ~~ Positive Use Cases}

One promising application of this technology lies in enhancing human-aligned vision models. By aligning machine perception more closely with human visual understanding, such models can achieve greater interpretability and robustness, particularly in complex visual environments. Another key domain is education and scientific visualization, where interpretable and structured visual understanding can aid in teaching abstract concepts, visualizing large datasets, or facilitating interactive learning tools.

\subsection*{F.2 ~~ Potential Risks}

Despite these benefits, there are notable risks. One major concern is the potential misuse of such models in adversarial scene manipulation, where an attacker might exploit a model's perceptual heuristics to deceive or mislead. Additionally, in sensitive fields like medical image analysis, overly rigid or misinterpreted model outputs could lead to diagnostic errors, especially if clinicians rely on model-generated insights without a thorough understanding of their limitations.

\subsection*{F.3 ~~ Safeguards}
We limit data release to PASCAL VOC and ImageNet subsets under their respective licenses. 

\section*{G ~~ Dataset Licensing and Attribution}

\subsection*{G.1 ~~ ImageNet and PASCAL VOC}
\begin{itemize}
    \item ImageNet: Licensed under ImageNet terms (\url{http://image-net.org/download})
    \item PASCAL VOC: CC BY 4.0 license
\end{itemize}

\subsection*{G.2 ~~ Codebase}
\begin{itemize}
    \item Based on PyTorch and timm library (Apache 2.0 License)
\end{itemize}

\section*{H ~~ Psychophysical Experiment Detail}
\begin{figure}[ht!]
     \centering
     \includegraphics[width=0.78\textwidth]{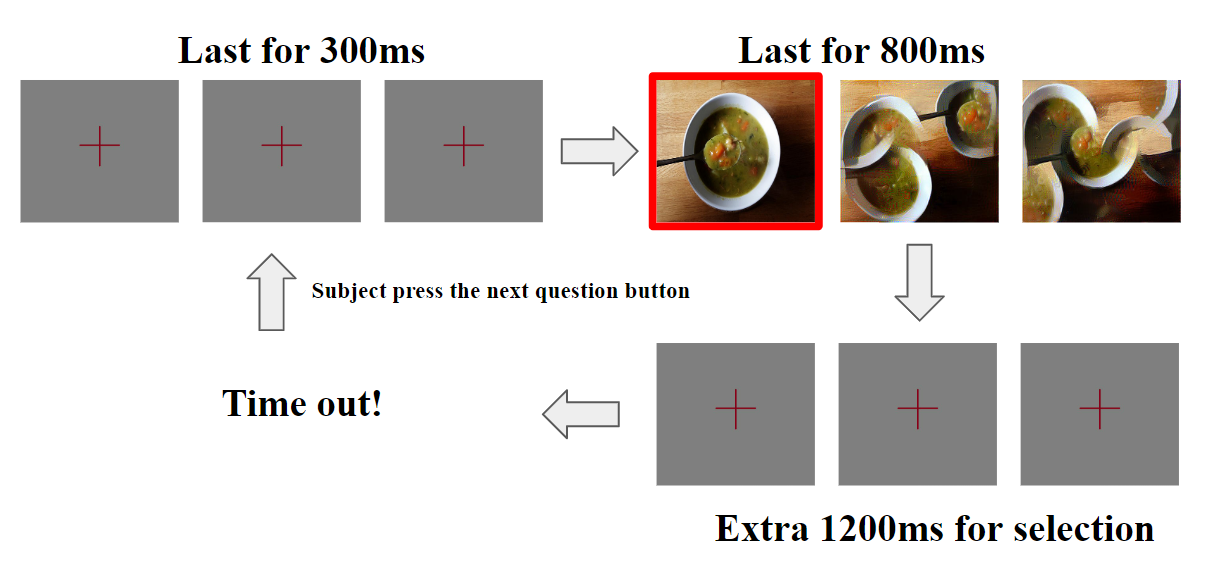} 
     \caption{Standard trial in the psychophysical experiment. Image in the red box is the correct answer.}
     \label{fig:normal_trail}
\end{figure}
\vspace{-2mm}

\begin{figure}[ht!]
     \centering
     \includegraphics[width=0.78\textwidth]{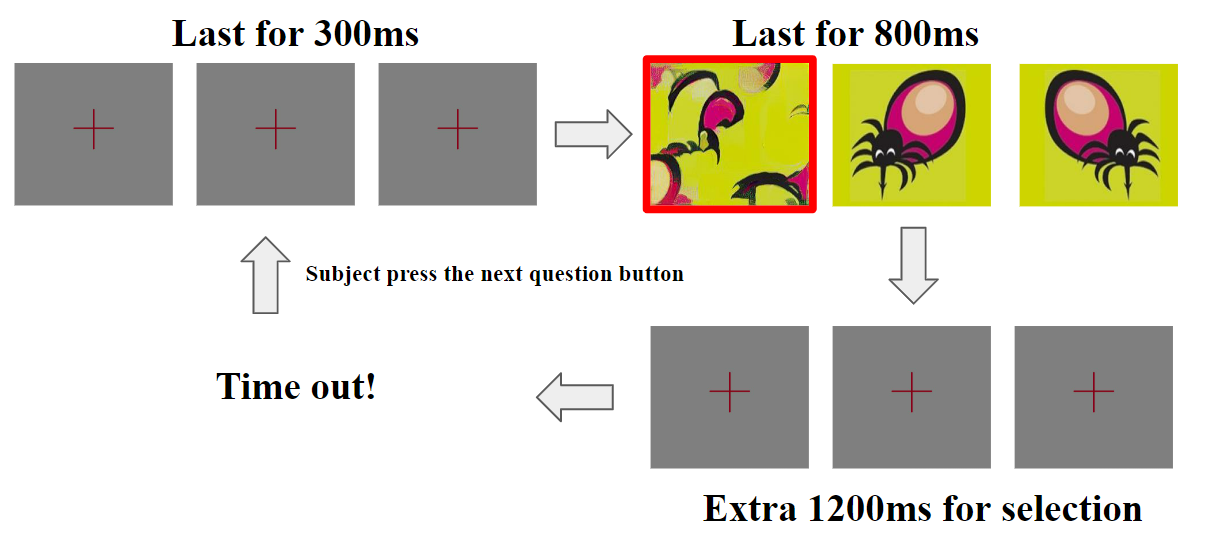} 
     \caption{Catch trial in the psychophysical experiment. Image in the red box is the correct answer}
     \label{fig:Catch_trial}
\end{figure}

Psychophysical experiments are conducted using a front-end web application developed in JavaScript. Subjects are instructed to "Find the image that is different from the other two" and can select their answers using keys `1', `2', or `3'. After making a selection, subjects press the spacebar to proceed to the next question.

The trial procedure is illustrated in Fig.\ref{fig:normal_trail}. A set of images appears on the screen after a 300 ms delay and remains visible for 800 ms. In a standard trial, two shape-distorted images and one original image are presented; the correct answer is the original image. Following the 800 ms display period, the images vanish, and subjects have an additional 1200 ms to make their selection, totaling 2 s for decision-making. If no selection is made within this time, the trial is marked as a timeout, and the response is considered invalid. Subjects are given the opportunity to take a break after every 100 images. To prevent the supervision signal, no feedback on answer correctness is provided during the test.

To mitigate the risk of the "oddity pop-out" test devolving into a mere "detection task"—where subjects might focus solely on identifying the original image rather than the one that differs—we incorporate extra catch trials into the experiments, as illustrated in Fig.\ref{fig:Catch_trial}.

One catch trial is presented after every 10 standard trials. In each catch trial, two "original images" are displayed: one is a mirrored version of the other, accompanied by a shape-distorted image. It is important to note that there is no overlap between the images used in catch trials and those used in standard trials. In these catch trials, the correct answer is actually the shape-distorted image. The rationale for incorporating such catch trials is to compel subjects to focus on identifying the "different" image rather than the "original" one, thereby aligning the task more closely with how deep learning models behave during DiST evaluation. Results from the catch trials are not included in the final performance metric.

\end{document}